\documentclass[conf]{new-aiaa}
\usepackage[utf8]{inputenc}

\usepackage{graphicx}
\usepackage{amsmath}
\usepackage[version=4]{mhchem}
\usepackage{siunitx}
\usepackage{longtable,tabularx}
\usepackage[bf]{subfigure}

\allowdisplaybreaks

\title{Total Least Squares for Optimal Pose Estimation}

\author{Saeed Maleki\footnote{Ph.D.~Candidate, Department of Mechanical and Aerospace Engineering, Email: saeedmal@buffalo.edu}
}
\affil{University at Buffalo, State University of New York, Amherst, NY, 14260-4400}
\author{Yang Cheng\footnote{Associate Professor, Department of Aerospace Engineering, Email: cheng@ae.msstate.edu.  Associate Fellow AIAA.}}
\affil{Mississippi State University, Mississippi State, MS, 39762}

\author{John Crassidis\footnote{SUNY Distinguished Professor, Samuel P.~Capen Chair Professor, Department of Mechanical and Aerospace Engineering, Email: johnc@buffalo.edu. Fellow AIAA.}}
\affil{University at Buffalo, State University of New York, Amherst, NY, 14260-4400}

\author{Matthias Schmid\footnote{Research Assistant Professor, Department of Automotive Engineering, Email: schmidm@clemson.edu.}}
\affil{Clemson University, Clemson, SC, 29634}

\begin{document}

\maketitle

\begin{abstract}
This work provides a theoretical framework for the pose estimation problem using total least squares for vector observations from landmark features. First, the optimization framework is formulated  with observation vectors extracted from point cloud features.
Then, error-covariance expressions are derived. The attitude and position solutions obtained via the derived optimization framework are proven to reach the bounds defined by the Cram\'er-Rao lower bound under the small-angle approximation of attitude errors.
The measurement data for the simulation of this problem is provided through a series of vector observation scans, and a fully populated observation noise-covariance matrix is assumed as the weight in the cost function to cover the most general case of the sensor uncertainty. Here, previous derivations are expanded for the pose estimation problem to include more generic correlations in the errors than previous cases involving an isotropic noise assumption. The proposed solution is simulated in a Monte-Carlo framework to validate the error-covariance analysis.
% The derivation is also verified in a Gazebo simulation environment in ROS to showcase the algorithm's efficacy.

\end{abstract}

% \section{Nomenclature}

% {\renewcommand\arraystretch{1.0}
% \noindent\begin{longtable*}{@{}l @{\quad=\quad} l@{}}
% $A$  & amplitude of oscillation \\
% $a$ &    cylinder diameter \\
% $C_p$& pressure coefficient \\
% $Cx$ & force coefficient in the \textit{x} direction \\
% $Cy$ & force coefficient in the \textit{y} direction \\
% c   & chord \\
% d$t$ & time step \\
% $Fx$ & $X$ component of the resultant pressure force acting on the vehicle \\
% $Fy$ & $Y$ component of the resultant pressure force acting on the vehicle \\
% $f, g$   & generic functions \\
% $h$  & height \\
% $i$  & time index during navigation \\
% $j$  & waypoint index \\
% $K$  & trailing-edge (TE) nondimensional angular deflection rate
% \end{longtable*}}

\section{Introduction}
In localization problems, such as simultaneous localization and mapping (SLAM), the part that determines the vehicle's orientation can be stated as an attitude estimation problem. The attitude estimation can be categorized in two ways. The first involves determining the orientation of the body frame with respect to a reference frame by taking multiple (at least two non-collinear) measurements in the form of vector observations from nearby features or the vehicle itself. The second involves filtering this observation information with models outlining the motion of the vehicle in a state estimation framework \cite{survey_nonlin_att}.
The advantage of the first is that an attitude solution is provided at each time-point in a deterministic manner without the transients or even divergence associated with filtering approaches. The advantage of the second is that attitude estimates can still be provided, even when one vector is available, and filtered estimates are given as part of the solution. One of the early derivations of the attitude determination for the first involved approaches that solve Wahba's problem \cite{wahba1965least}, some of which have closed-form solutions  \cite{markley2000quaternion}. Wahba's problem also has connections to the attitude and position determination problem \cite{eggert1997estimating}, which is called pose estimation. This pose estimation problem, which involves both attitude and position estimation, is addressed in this paper.

Pose estimation based on imagery data has various applications for human-computer interaction in virtual environments and augmented reality systems, in which monocular or stereo cameras are used for hand pose estimation \cite{ErolAli2007Vhpe} with several degrees of freedom. Image-based relative pose determination is categorized in two classes of general algorithms \cite{kelsey2006vision}:
model-based and non-model-based techniques. Model-based solutions use an initial known model of the object that includes dimensions, shape or texture, and matches the visual features to determine the model's unknown parameters.
Template matching \cite{jurie2002real} and contour tracking \cite{isard1998condensation} are examples for this class of pose estimation solutions. Non-model-based solutions do not employ a model of the object but rather determine the pose from a sequence of images such as structure from motion problem \cite{schonberger2016structure}. Deep-learning approaches are also another series of solutions for relative pose estimation from image data sets \cite{melekhov2017relative, li2018deepim, toshev2014deeppose}.

There is another category of pose estimation solutions that rely on Lidar data sets, which use point cloud data structures \cite{rusu20113d} to illustrate the objects in the environment. Reference \cite{opromolla2015uncooperative} provides a solution for Lidar-based pose estimation for non-cooperative objects. The iterative closest point (ICP) approach \cite{rusinkiewicz2001efficient} is one of the well-known approaches to match the data structures from different Lidar frames, which is an essential problem for the alignment of 3D point cloud data from different scans. Segmentation of the object of interest from its surroundings is a crucial task in point cloud pre-processing \cite{woo2002new}. Connected components \cite{trevor2013efficient} and super-point graphs \cite{landrieu2018large} are two examples of such methods. Feature extraction is another primary step to provide the necessary information from the point cloud segmented objects. A class of stable algorithms exist for this purpose, such as SIFT \cite{lindeberg2012scale} and SURF \cite{bay2006surf}. Feature matching and correspondence of different features in several Lidar scans is another issue associated with SLAM applications. One solution for the SLAM problem involves matching the histograms extracted from the features  \cite{rusu2008aligning}. Reference \cite{serafin2015nicp} utilizes a point as well as the local properties of its neighborhood, e.g.~the curvature and normal direction, to align point clouds.  

It is assumed in this paper that the segmentation, feature extraction, and feature matching steps for data sets are already solved. The focus of the present work is to solve the pose estimation from matched features as vector observations. The pose estimation problem from observation vectors is well-studied \cite{hashim2020attitude}. However, little work exists on efficiently determining the associated error-covariance expressions, i.e.~ones that achieve the Cram\'er-Rao lower bound \cite{alma9938879548204803}.
Reference \cite{cheng2021optimal} derives efficient error-covariance expressions, but assumes isotropic matrices for the vector-measurement covariance. %or just scalars \cite{haralick1989pose}. It is shown in \cite{cheng2021optimal} that the optimal pose estimation problem is related to total least squares (TLS), which includes errors in both the ``design matrix'' and measurement observation \cite{crassidis2019maximum}.
In this paper, the problem of the most general case of sensor uncertainty that includes fully-populated measurement covariance matrices, which also account for the correlations between the measurement vectors, is solved. Furthermore, error-covariance expressions are derived that achieve the Cram\'er-Rao lower bound, and reduce to the expressions shown in \cite{cheng2021optimal} when the measurement covariance matrices are isotropic. An example that can be benefited from the analysis shown in this paper involves medical image databases \cite{estepar2004robust}.

\begin{figure}
  \centering
  \includegraphics[width=3in]{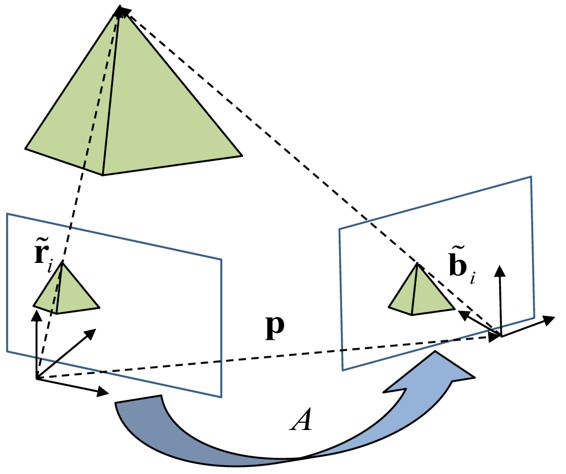}\\
  \caption{Geometric interpretation of the pose estimation problem. Image from \cite{alma9938879548204803}}\label{fig:geometry_calibration}
\end{figure}

\section{Problem Statement}
The pose estimation problem is described as finding the pose of a sensor attached to a vehicle with respect to another sensor or a reference frame.
The pose itself consists of two components: 1) a translation vector that connects the center of the two frames, and 2) an attitude matrix for the relative orientation of the unit vectors of the coordinate systems.
A geometric model relating the pose of the reference frame to the body frame is shown in Figure \ref{fig:geometry_calibration}.
The associated geometric constraint acts as a measurement model.
The vectors ${\tilde{\mathbf r}_i}$ and ${\tilde{\mathbf b}_i}$ are the coordinates of a point of interest, i.e.~an observation feature of a landmark in the environment, with respect to the reference and body frames, respectively, and $\mathbf{p}$ refers to the translation vector between the two coordinates.

The estimation approach provides an estimate of the attitude as well as the translation vector based on the measurement information, denoted by ${\tilde{\mathbf r}_i}$ and ${\tilde{\mathbf b}_i}$.
The estimate error, which comprises the difference between the transformed version of ${\tilde{\mathbf r}_i}$ and ${\tilde{\mathbf b}_i}$ is given by
\begin{align}
    \mathbf{e}_i(\hat{A},\hat{\mathbf{p}})={\tilde{\mathbf b}_i}-\hat{A}{\tilde{\mathbf r}_i}+{\hat{\mathbf p}}
\end{align}
where $\hat{A}$ is the estimated attitude matrix, and $\hat{\mathbf{p}}$ is the estimated translation vector.
In case of perfect measurements, the error samples $\mathbf{e}_i$ are all zero, and the problem can be solved with the measurement model of the form
\begin{equation}\label{perfect_measurement}
    \mathbf{b}_i=\hat{A}\mathbf{r}_i-\hat{\mathbf{p}}
\end{equation}
where $\mathbf{b}_i$ and $\mathbf{r}_i$ denote the true values of the observation vectors.
But in an actual applications, these errors are not zero, which leads to an optimization problem derived from a constrained maximum likelihood approach \cite{alma9938879548204803}, given by
\begin{equation}
\begin{gathered}
    \underset{\hat{A},{\hat{\mathbf p}}}{\min}\text{\ }J=\frac{1}{2}\sum_{i=1}^n\left({\tilde{\mathbf b}_i}-\hat{A}{\tilde{\mathbf r}_i}+\hat{\mathbf{p}}\right)^T R_i^{-1}\left({\tilde{\mathbf b}_i}-\hat{A}{\tilde{\mathbf r}_i}+{\hat{\mathbf p}}\right)\label{eqn:cost_att}\\
    \text{subject}\text{\ }\text{to:\ }\hat{A}^T \hat{A}= I_{3\times3} \text{\ },\text{\ }\det(\hat{A})=1
\end{gathered}
\end{equation}
where $I_{3\times3}$ is an identity matrix, and $R_i$ is the measurement covariance matrix that accounts for errors on both $\tilde{\mathbf b}_i$ and $\tilde{\mathbf r}_i$, as well as correlations that exist between them.
The determinant condition is required so that $\hat{A}$ is a proper attitude matrix. % in which there is a weight matrix $W$ to showcase the level of noise we expect for each dimension in the error signal $\mathbf{e_i}$.
The observation errors are defined as
\begin{subequations}
\begin{gather}
    \boldsymbol{\Delta} \mathbf{b}_i={\tilde{\mathbf b}}_i-\mathbf{b}_i\label{eqn:def_Delta_b}\\
    \boldsymbol{\Delta} \mathbf{r}_i={\tilde{\mathbf r}}_i-\mathbf{r}_i\label{eqn:def_Delta_r}
\end{gather}
\end{subequations}
%in which $\mathbf{b_i}$ and $\mathbf{r_i}$ denote the true values of the observation vectors. 
It is assumed that zero-mean Gaussian measurement errors exist, with 
\begin{subequations}
\begin{align}
    &\mathcal{E}\{\boldsymbol{\Delta} \mathbf{b}_i\boldsymbol{\Delta} \mathbf{b}^T_i\}=R_{b_i}\label{eqn:cov_noise_bi}\\
    &\mathcal{E}\{\boldsymbol{\Delta} \mathbf{r}_i\boldsymbol{\Delta} \mathbf{r}^T_i\}=R_{r_i}\label{eqn:cov_noise_ri}\\
    &\mathcal{E}\{\boldsymbol{\Delta} \mathbf{r}_i\boldsymbol{\Delta} \mathbf{b}^T_i\}=\mathcal{E}\{\boldsymbol{\Delta} \mathbf{b}_i\boldsymbol{\Delta} \mathbf{r}^T_i\}^T=R_{rb_i}\label{eqn:cov_noise_rbi}\\
    &R_i=\mathcal{E}\left\{\begin{bmatrix}\boldsymbol{\Delta} \mathbf{r}_i\\\boldsymbol{\Delta} \mathbf{b}_i\end{bmatrix}\begin{bmatrix}\boldsymbol{\Delta} \mathbf{r}^T_i&\boldsymbol{\Delta} \mathbf{b}^T_i\end{bmatrix}\right\}=\begin{bmatrix}R_{r_i}&R_{rb_i}\\R^T_{rb_i}& R_{b_i}\end{bmatrix}\label{eqn:R_i}
\end{align}
\end{subequations}
where $\mathcal{E}$ denotes expectation. The optimization problem in Eq.~\eqref{eqn:cost_att} can be shown to be related to a total least squares (TLS) problem  \cite{cheng2021optimal}. 
%The optimization problem for pose estimation results in the full pose of the body frame with respect to the reference frame.
%As we see in figure \ref{fig:geometry_calibration}, the measurement vectors as well as the attitude matrix and the translation vectors are erroneous and because of the linear structure of the constraints in Eq. (\ref{perfect_measurement}), we can use total least squares(TLS) as a solution to this problem \cite{qing2018weighted}.

This paper solves the pose estimation problem in Eq.~\eqref{eqn:cost_att} with the fully populated noise covariance matrices $R_{r_i}$, $R_{b_i}$ and $R_{{rb}_i}$ using TLS.
Note that although there are nine components in the attitude matrix $A$, only three of them are independent, so the attitude estimation solution can be accomplished with a minimum of three parameters, which can be Euler angles or any other minimal attitude parameterization \cite{shuster1993survey}. 
%of rotation from the reference to the current body frame which is summarized 
Regardless of the attitude parameterization used in the optimization process, the estimated attitude can be related to the true attitude using an attitude error-vector involving small roll, pitch and yaw angles, as will be seen in section \ref{tls_derivation}.

The derivation begins with an introductory formulation of TLS in section \ref{intro_tls}.
A cost function is developed in section \ref{tls_derivation} based on the pose estimation problem from Eq.~\eqref{eqn:cost_att}.
The necessary conditions for efficient pose estimation are derived, and the cost function will be written in the attitude-only format for the sake of simplicity in the proceeding derivations.
A linear approximation of the attitude error is then derived to approximate the cost function within second-order terms in the attitude errors.
The efficient estimations for the attitude and translation vectors are used to derive the error-covariance of the estimate errors and residuals in section \ref{cov_analysis}, which are beneficial for control purposes in a sense that the most accurate pose estimates are provided to the control logic to prevent extra effort.
% Section (\ref{pseudo_code}) provides the logic steps to solve the problem numerically and 
Finally, a numerical verification of the proposed TLS estimator is shown in the context of a Monte-Carlo simulation in section \ref{sec_monte_carlo}.
% as well as a Gazebo Lidar simulation in ROS
% in sections (\ref{sec_monte_carlo}) and (\ref{sec_gazebo}) respectively.

\subsection{Overview of Linear Least Squares and Total Least Squares}\label{intro_tls}
This section briefly introduces linear and total least squares, how they are related, and their differences.
For a more in-depth review of the TLS, the reader is referred to \cite{golub1973some, markovsky2007overview, golub1980analysis}. 
%In the estimation problems, the equation that relates the sensor data to the unknowns is called the $\textit{sensor model}$ or $\textit{measurement model}$.
Consider the measurement model of the form 
\begin{equation}\label{eqn:meas_model_LS}
    \tilde{\mathbf{y}}=H\mathbf{x}+\boldsymbol{\Delta}\mathbf{y}
\end{equation}
where $H $ is an $m\times n$ deterministic matrix with no errors, $\mathbf{x}$ is the $n\times 1$ vector of unknowns, $\tilde{\mathbf{y}}$ is the $m \times 1$ measurement vector, and $\boldsymbol{\Delta}\mathbf{y}$ is the $m \times 1$ measurement error-vector, % at each sample of the measurement data $\tilde{\mathbf{y}}_{m\times 1}$.
The least squares estimate of $\mathbf{x}$ is given by solving the following problem:
\begin{equation}\label{cost_ls}
\begin{gathered}
    \underset{\hat{\mathbf{x}}}{\min}\text{\ }J=\frac{1}{2}\boldsymbol{\Delta}\mathbf{y}^T\boldsymbol{\Delta}\mathbf{y} \\
    \text{subject to: }\mathbf{\hat{y}}=H\hat{\mathbf{x}}
\end{gathered}
\end{equation}
where the number of measurement samples stacked vertically in $\tilde{\mathbf{y}}$ should be more than the number of unknowns for the problem to be observable.
The main underlying assumption in the statistical analysis of least squares is that $\tilde{\mathbf{y}}$ has a Gaussian distribution with the conditional likelihood function given by
\begin{equation}\label{eqn:liklihood_ls}
    p(\mathbf{\tilde{y}}|\mathbf{x})=\frac{1}{(2\pi)^{\frac{m}{2}}\big[\det(R_{yy})\big]^{\frac{1}{2}}}\text{exp}\left\{-\frac{1}{2}(\mathbf{\tilde{y}}-H \mathbf{x})^TR^{-1}_{yy}(\mathbf{\tilde{y}}-H \mathbf{x})\right\}
\end{equation}
where the distribution mean is denoted by $H\mathbf{x}$ and the covariance is $R_{yy}$. 
Because of the properties of the exponential function, maximizing the likelihood function \ref{eqn:liklihood_ls} is equivalent to minimizing the negative of the log-likelihood. % which is the original cost function in \ref{cost_ls}.
The mean and error-covariance of the estimate are given by
\begin{subequations}
\begin{align}
    \mathcal{E}\{\hat{\mathbf{x}}\}&=\left(H^T R^{-1}_{yy} H\right)^{-1}H^TR^{-1}_{yy}\tilde{\mathbf{y}}\\
    \text{cov}\{\hat{\mathbf{x}}\}&=\left(H^T R^{-1}_{yy} H\right)^{-1}
\end{align}
\end{subequations}
which shows that the least squares estimate is unbiased. %are the only couple of distribution moments we need to completely specify the Gaussian structure of the unknown $\mathbf{x}$.

As stated previously, the design matrix $H$ in the least squares measurement model in Eq.~\eqref{eqn:meas_model_LS} has no errors.
If this underlying assumption does not exist anymore, which happens in many applications, as will be seen in the SLAM problem in section \ref{tls_derivation}, then another formulation must be used to consider the errors in the design matrix, which leads to the TLS problem, with paramaters defined by
%Total Least Squares (TLS) is a solution for this problem with the measurement model as 
\begin{subequations}
\begin{gather}
  \tilde{\mathbf y}=\mathbf{y}+\boldsymbol{\Delta}\mathbf{y}\\
  \mathbf{y}=H\mathbf{x}\\
  \tilde{H}=H+\Delta{H}
\end{gather}
\end{subequations}
where  $\Delta H$ shows the errors in the design matrix. Consider the following augmented matrix:
\begin{equation}
    D=\begin{bmatrix}H & \mathbf{y}\end{bmatrix}
\end{equation}
The conditional likelihood function of the TLS problem is defined by 
\begin{equation}
     p(\tilde{D}|D)=\frac{1}{(2\pi)^{\frac{m}{2}}\big[\det(R)\big]^{\frac{1}{2}}}\text{exp}\left\{-\frac{1}{2}\text{vec}(\tilde{D}-D)^TR^{-1}\text{vec}(\tilde{D}-D)\right\}
\end{equation}
where vec operator stacks all columns of a matrix in a single column.
The maximum likelihood approach for this cost function leads to the minimization of the log-likelihood function as
\begin{equation}\label{eqn:cost_TLS_D}
\begin{gathered}
    J(\hat{D})=\frac{1}{2}\text{vec}(\tilde{D}-\hat{D})^T R^{-1} \text{vec}(\tilde{D}-\hat{D})\\
    \text{subject to}: \hat{D}\,{\hat{\mathbf z}}=\textbf{0} \\
%    \hat{\mathbf{z}}=\begin{bmatrix}^T & -1\end{bmatrix}^T
\end{gathered}
\end{equation}
where $\hat{\mathbf{z}}= [{\hat{\mathbf x}}^T \,\,\, -1]^T$. A unique solution for this problem can be obtained if $\text{rank}(D)=n$.
Also, $R$ is the covariance matrix that accounts for the errors in both $\tilde{\mathbf y}$ and $\tilde{H}$. % in which $n$ is the size of the unknown vector $\mathbf{x}$ and $R$ shows the weight or coefficient matrix of the cost function which is a function of the measurement noise covariance as well as the unknown vector $\mathbf{x}$.
Although the TLS solution is known to be biased, the TLS problem is proven to reach the Cram\'er-Rao lower bound (CRLB) \cite{cramer1999mathematical} for the estimate error-covariance to within first-order error-terms, and therefore is an {\it efficient} estimator \cite{crassidis2019maximum}.
Closed-form solutions for the TLS problem are possible only when $R$ is an isotropic matrix.

\subsection{Pose Estimation Sensor Model}\label{p_est_sensor_model}
%We begin the derivation by the geometry of the SLAM problem illustrated in figure (\ref{fig:geometry_calibration}) in which a sensor takes scans an object and the observation vector to a selected feature appears as $\mathbf{\tilde{b}_i}$ and $\mathbf{\tilde{r}_i}$ in body and reference frames respectively.
%The $\tilde{.}$ sign stands for the noisy measurement vectors.
The relation between the true vectors $\mathbf{b}_i$ and $\mathbf{r}_i$ is given by
\begin{equation}
\begin{split}
    \mathbf{b}_i&=A\mathbf{r}_i-\mathbf{p} \\
    &=\begin{bmatrix}\mathbf{r}^T_i & \mathbf{0}_{1\times3} & \mathbf{0}_{1\times3}\\
    \mathbf{0}_{1\times3} & \mathbf{r}^T_i & \mathbf{0}_{1\times3} \\
    \mathbf{0}_{1\times3} & \mathbf{0}_{1\times3} & \mathbf{r}^T_i  \end{bmatrix}\begin{bmatrix}\mathbf{a}_1\\
    \mathbf{a}_2\\
    \mathbf{a}_3\end{bmatrix}-\mathbf{p}\\
    &=H_i\mathbf{x}-\mathbf{p} \\ \label{eqn:det_tls_sensor_model}
    &\equiv\mathbf{y}_i
\end{split}
\end{equation}
in which Eq.~\eqref{eqn:kronecker_vec} has been used. The matrix $H_i$ is the individual sensor model design matrix and $\mathbf{a}_i\ ,\  i=1,\,2,\,3$, are the columns of the attitude matrix $A$.
However, the perfect measurement model is not realistic because of noise in the design matrix as well as the observation vectors of Eq.~\eqref{eqn:det_tls_sensor_model}.
Therefore, in the actual version of the sensor model in Eq.~\eqref{eqn:det_tls_sensor_model}, the following relation is used:
\begin{equation}
\begin{split}
    {\tilde{\mathbf b}_i}-\boldsymbol{\Delta}\mathbf{b}_i&=A({\tilde{\mathbf r}_i}-\boldsymbol{\Delta}\mathbf{r}_i)-\mathbf{p}\\
    &=(\tilde{H}_i-\Delta H_i){\mathbf x}-\mathbf p \\
    &\equiv{\tilde{\mathbf y}_i}-\boldsymbol{\Delta} \mathbf{y}_i
\end{split}
\end{equation}
where the design matrix $\tilde{H}_i$ and the observation vector $\tilde{\mathbf{y}}_i$ have the errors of $\Delta H_i$ and $\boldsymbol{\Delta} \mathbf{y}_i$, respectively.
Because the model is linear in terms of the unknowns $\mathbf{x}$ and the translation vector $\mathbf{p}$, then the problem can be posed using a TLS formulation with the constraint 
\begin{equation}
 {\hat{\mathbf b}_i}=\hat{A}{\hat{\mathbf r}_i}-{\hat{\mathbf p}}
\end{equation}
which is equivalent to
\begin{equation}\label{eqn:dz=0}
\hat{D}_i{\hat{\mathbf z}}-{\hat{\mathbf p}}=\mathbf{0}    
\end{equation}
where $\hat{\mathbf{z}}= [{\hat{\mathbf x}}^T \,\,\, -1]^T$ and $\hat{D}_i=[\hat{H}_i  \,\,\, \mathbf{\hat{y}}_i]$.
%with definition of $\hat{\mathbf{z}}=\begin{bmatrix}\hat{\mathbf{x}}^T & -1\end{bmatrix}^T$ and $\hat{\mathbf{x}}$ being the columns of the attitude estimate stacked together in a $9\times 1$ vector.

\subsection{Total Least Squares Derivation for Pose Determination}\label{tls_derivation}
The TLS cost function is given by
\begin{equation}\label{eqn:original_J_tls}
    J(\hat{D}_i)=\frac{1}{2}\sum_{i=1}^n\text{vec}(\tilde{D}_i-\hat{D}_i)^T R_{D_i}^{-1} \text{vec}(\tilde{D}_i-\hat{D}_i)
\end{equation}
where $R_{D_i}$ is the covariance of $\text{vec}(\tilde{D}_i)$.
The augmented cost function that includes the linear constraint in Eq.~\eqref{eqn:dz=0} is given by
\begin{equation}
\begin{split}
    J_a(\hat{D}_i,\boldsymbol{\lambda}_i)&=\frac{1}{2}\sum_{i=1}^n\text{vec}(\tilde{D}_i-\hat{D}_i)^T R_{D_i}^{-1} \text{vec}(\tilde{D}_i-\hat{D}_i)+\boldsymbol{\lambda}_i^T\left(\hat{D}_i\hat{\mathbf{z}}-\hat{\mathbf{p}}\right)
\end{split}
\end{equation}
where $n$ is the number of features in each sensor scan. Taking the partial derivative of the augmented cost function with respect to $\text{vec}(\hat{D}_i)$ and utilizing Eq.~\eqref{eqn:kronecker_vec} gives
\begin{equation}
    \frac{\partial J_a}{\partial \text{vec}(\hat{D}_i)}=-R^{-1}_{D_i}\text{vec}(\tilde{D}_i-\hat{D}_i)+(\hat{\mathbf{z}}^T\otimes I_{3\times 3})^T\boldsymbol{\lambda}_i
\end{equation}
where $\otimes$ is the Kronecker product \cite{steeb2011matrix}.
From the necessary condition, this partial derivative should be a zero vector, and therefore
\begin{equation}\label{eqn:vecdhat}
    \text{vec}(\hat{D}_i)=\text{vec}(\tilde{D}_i)-R_{D_i}(\hat{\mathbf{z}}^T\otimes I_{3\times 3})^T\boldsymbol{\lambda}_i
\end{equation}
Using Eq.~\eqref{eqn:kronecker_vec}, the constraint in Eq.~\eqref{eqn:dz=0} can be written as
\begin{equation}\label{eqn:kron_constraint}
    (\hat{\mathbf{z}}^T\otimes I_{3\times 3})\text{vec}(\hat{D}_i)-\hat{\mathbf{p}}=\mathbf{0}
\end{equation}
Substituting Eq.~\eqref{eqn:vecdhat} into the constraint gives %results in \eqref{eqn:kron_constraint} 
\begin{equation}
    (\hat{\mathbf{z}}^T\otimes I_{3\times 3})\left[\text{vec}(\tilde{D}_i)-R_{D_i}(\hat{\mathbf{z}}^T\otimes I_{3\times 3})^T\boldsymbol{\lambda}_i\right]-\hat{\mathbf{p}}=\mathbf{0}
\end{equation}
Solving for $\boldsymbol{\lambda_i}$ leads to
\begin{equation}\label{eqn:solvelambda}
    \boldsymbol{\lambda}_i=Q^{-1}_{\hat{\lambda}_i}\left[ (\hat{\mathbf{z}}^T\otimes I_{3\times 3})\text{vec}(\tilde{D}_i)-\hat{\mathbf{p}}\right]
\end{equation}
in which $Q_{\hat{\lambda}_i}=(\hat{\mathbf{z}}^T\otimes I_{3\times 3})R_{D_i}(\hat{\mathbf{z}}^T\otimes I_{3\times 3})^T$.
Note that in this paper, $Q_{\hat{\lambda}_i}$ is considered to be a positive-definite matrix, though it might be singular in some sensor models, such as the Quest Measurement Model \cite{shuster1981three}. 
This problem can be solved using a similar approach to the eigenvalue decomposition in \cite{amiri2017weighted}.
Using the necessary condition for $\hat{D}_i$ from Eq.~\eqref{eqn:vecdhat}, and substituting the Lagrangian multiplier $\boldsymbol{\lambda}_i$ from Eq. \eqref{eqn:solvelambda} gives
\begin{equation}
    \begin{split}
        \text{vec}(\tilde{D}_i)-\text{vec}(\hat{D}_i)&=R_{D_i}(\hat{\mathbf{z}}^T\otimes I_{3\times 3})^TQ^{-1}_{\hat{\lambda}_i}\left[ (\hat{\mathbf{z}}^T\otimes I_{3\times 3})\text{vec}(\tilde{D}_i)-\hat{\mathbf{p}}\right]\\
    &=R_{D_i}(\hat{\mathbf{z}}^T\otimes I_{3\times 3})^TQ^{-1}_{\hat{\lambda}_i}\left(\tilde{D}_i\hat{\mathbf{z}}-\hat{\mathbf{p}}\right)\\
    &=-R_{D_i}(\hat{\mathbf{z}}^T\otimes I_{3\times 3})^TQ^{-1}_{\hat{\lambda}_i}\left(\hat{\mathbf{p}}-\tilde{D}_i\hat{\mathbf{z}}\right)\label{eqn:step2vecdd}
    \end{split}
\end{equation}
 
Substituting Eq.~\eqref{eqn:step2vecdd} into the original cost function in Eq.~\eqref{eqn:original_J_tls} yields 
\begin{equation}
    J(\hat{\mathbf{z}},\hat{\mathbf{p}})=\frac{1}{2}\sum_{i=1}^n\left(\hat{\mathbf{p}}-\tilde{D}_i\hat{\mathbf{z}}\right)^T Q^{-1}_{\hat{\lambda}_i}\left(\hat{\mathbf{p}}-\tilde{D}_i\hat{\mathbf{z}}\right)
\end{equation}
This results in a formulation, satisfying the constraint in Eq.~\eqref{eqn:dz=0} in the cost function in Eq.~\eqref{eqn:original_J_tls}. Then in terms of the observation vectors $\tilde{\mathbf{b}}_i$ and $\tilde{\mathbf{r}}_i$, and from $\hat{\mathbf{p}}-\tilde{D}_i\hat{\mathbf{z}}=\tilde{\mathbf{b}}_i-\hat{A}\tilde{\mathbf{r}}_i+\hat{\mathbf{p}}$, the following expression is given:
\begin{equation}\label{eqn:cost_ahat_phat}
    J(\hat{A},\hat{\mathbf{p}})=
    \frac{1}{2}\sum_{i=1}^n\left(\tilde{\mathbf{b}}_i-\hat{A}\tilde{\mathbf{r}}_i+\hat{\mathbf{p}}\right)^T Q^{-1}_{\hat{\lambda}_i}\left(\tilde{\mathbf{b}}_i-\hat{A}\tilde{\mathbf{r}}_i+\hat{\mathbf{p}}\right)
\end{equation}
For simplicity in the proceeding derivations, the cost function can be written only in terms of the attitude matrix, which is known as the attitude-only cost function. For this purpose,  $\hat{\mathbf{p}}$ needs to be eliminated from the cost function in Eq.~\eqref{eqn:cost_ahat_phat}. The necessary condition for the translation vector $\hat{\mathbf{p}}$ results in
\begin{equation}\label{eqn:def_phat}
    \hat{\mathbf{p}}=-\left(\sum_{i=1}^{n}Q^{-1}_{\hat{\lambda_i}}\right)^{-1} \left[\sum_{i=1}^{n}Q^{-1}_{\hat{\lambda_i}}\left(\tilde{\mathbf{b}}_i-\hat{A}\tilde{\mathbf{r}}_i\right)\right]
\end{equation}
Substituting the optimal value of $\hat{\mathbf{p}}$ into Eq.~\eqref{eqn:cost_ahat_phat}, the attitude-only cost function becomes
\begin{equation}\label{eqn:cost_ahat}
    J(\hat{A})=\frac{1}{2}\Bigg[\sum_{i=1}^{n}\big(\tilde{\mathbf{b}}_i-\hat{A}\tilde{\mathbf{r}}_i\big)^T Q^{-1}_{\hat{\lambda_i}}\big(\tilde{\mathbf{b}}_i-\hat{A}\tilde{\mathbf{r}}_i\big)\Bigg]-\frac{1}{2}\Bigg[\sum_{i=1}^{n}Q^{-1}_{\hat{\lambda_i}}\big(\tilde{\mathbf{b}}_i-\hat{A}\tilde{\mathbf{r}}_i\big)\Bigg]^T\Bar{Q}_{\hat{\lambda}}\Bigg[\sum_{i=1}^{n}Q^{-1}_{\hat{\lambda_i}}\big(\tilde{\mathbf{b}}_i-\hat{A}\tilde{\mathbf{r}}_i\big)\Bigg]
\end{equation}
in which
\begin{subequations}
    \begin{gather}
        \mathbf{\Tilde{b}}_i=\mathbf{b}_i+\boldsymbol{\Delta} \mathbf{b}_i\label{eqn:b_tilde_def}\\
    \mathbf{\Tilde{r}}_i=\mathbf{r}_i+\boldsymbol{\Delta} \mathbf{r}_i\label{eqn:r_tilde_def}
    \end{gather}
\end{subequations}
\begin{equation}
    \Bar{Q}_{\hat{\lambda}}=\left(\sum_{i=1}^{n}Q^{-1}_{\hat{\lambda_i}}\right)^{-1}\label{q_cap_bar}
\end{equation}
are the observation vectors with their corresponding covariance shown in Eq.~\eqref{eqn:cov_noise_bi}, \eqref{eqn:cov_noise_ri} and \eqref{eqn:cov_noise_rbi}. Also it is proven in \cite{cheng2021optimal} that the weight matrix $Q_{\hat{\lambda}_i}$ can be derived as a function of the attitude matrix and the observation noise covariance matrices as
\begin{equation}\label{eqn:q_lam_hat_i}
    Q_{\hat{\lambda}_i}=\hat{A}R_{r_i}\hat{A}^T-\hat{A}R_{rb_i}-R_{rb_i}^T\hat{A}^T+R_{b_i}
\end{equation}
 
\subsection{Covariance Analysis of the Estimates and Residuals}\label{cov_analysis}
The covariance expressions for the attitude as well as the translation and observation vector estimates are now derived.
Note that the cost function can be written in terms of the attitude error $\boldsymbol{\delta} \boldsymbol{\alpha}$, since only 3 independent components exist inside of the attitude matrix. The relation between the true and the estimated attitude matrix can be expressed as
\begin{equation}
    \hat{A}=\text{exp}(-[\boldsymbol{\delta} \boldsymbol{\alpha}\times])A
\end{equation}
where $[.\times]$ denotes the cross product matrix of a vector \cite{markley2014fundamentals}.
Using a small angle assumption and a first-order approximation of the attitude error, the attitude estimate can be written as
\begin{equation}
    \hat{A}\approx \big(I_{3\times3}-[\boldsymbol{\delta} \boldsymbol{\alpha}\times ]\big)A\label{eqn:att_error_1st}
\end{equation}
The cost function needs be derived up to second-order in terms of attitude error $\boldsymbol{\delta} \boldsymbol{\alpha}$, since  covariance expressions for a first-order approximation of the unknown errors are sought. The derivation begins with the approximation of the error-terms inside the cost function of Eq.~\eqref{eqn:cost_ahat}. The attitude approximation in Eq.~\eqref{eqn:att_error_1st}, Eqs.~\eqref{eqn:kronecker_vec} and \eqref{eqn:cross_prod_def} from the appendix, as well as Eqs.~\eqref{eqn:b_tilde_def} and \eqref{eqn:r_tilde_def}, are utilized for the formulation of the observation vectors, yielding 
\begin{equation}
\begin{split}
    \tilde{\mathbf{b}}_i-\hat{A}\tilde{\mathbf{r}}_i&=\mathbf{b}_i+\boldsymbol{\Delta}\mathbf{b}_i-\big(I_{3\times3}-[\boldsymbol{\delta} \boldsymbol{\alpha}\times]\big)A(\mathbf{r}_i+\boldsymbol{\Delta}\mathbf{r}_i)\\
    &\approx \mathbf{b}_i-A\mathbf{r}_i+\boldsymbol{\Delta} \mathbf{b}_i-A\boldsymbol{\Delta} \mathbf{r}_i - [A\mathbf{r}_i\times]\boldsymbol{\delta} \boldsymbol{\alpha}
\end{split}
\end{equation}
The following abbreviations are introduced for simplicity:
\begin{subequations}
    \begin{align}
        \boldsymbol{\Delta} \mathbf{a}_i&\equiv\boldsymbol{\Delta} \mathbf{b}_i-A\boldsymbol{\Delta} \mathbf{r}_i\label{eqn:def_Dai}\\
    \mathbf{p}&=A\mathbf{r}_i-\mathbf{b}_i\\
    \boldsymbol \nu_i &\equiv-\boldsymbol{\Delta} \mathbf{a}_i+\mathbf{p}\label{eqn:def_nu_i}\\
    \mathcal{A}_i&\equiv[A\mathbf{r}_i\times]
    \end{align}
\end{subequations}
This allows for the reformulation of the error-term $\tilde{\mathbf{b}}_i-\hat{A}\tilde{\mathbf{r}}_i$ as
\begin{align}
    \label{bti-ati}
    \tilde{\mathbf{b}}_i-\hat{A}\tilde{\mathbf{r}}_i\approx -\boldsymbol \nu_i-\mathcal{A}_i\boldsymbol{\delta} \boldsymbol{\alpha}
\end{align}
Note that $Q_{\hat{\lambda}_i}$ in Eq.~\eqref{eqn:q_lam_hat_i} is also a function of the attitude estimate $\hat{A}$ and subsequently of the attitude error $\boldsymbol{\delta} \boldsymbol{\alpha}$. As the neighboring terms are already a function of the first-order attitude error, any other term besides the attitude-error dependent ones in $Q_{\hat{\lambda}_i}$ (terms that are independent of $\boldsymbol{\delta} \boldsymbol{\alpha}$) are not kept for the second-order approximation of the cost function. The matrix $Q_{\hat{\lambda}_i}$ is now written as
\begin{equation}
    \begin{split}
        \label{eqn:expand_Q}
    Q_{\hat{\lambda}_i}&=\hat{A}R_{r_i}\hat{A}^T-\hat{A}R_{rb_i}-R_{rb_i}^T\hat{A}^T+R_{b_i} \\
    &=\big(I_{3\times3}-[\boldsymbol{\delta} \boldsymbol{\alpha}\times]\big)A R_{r_i} A^T \big(I_{3\times3}-[\boldsymbol{\delta} \boldsymbol{\alpha}\times]\big)^T -\big(I_{3\times3}-[\boldsymbol{\delta} \boldsymbol{\alpha}\times]\big)A R_{rb_i}-R_{rb_i}^T A^T (I_{3\times3}-[\boldsymbol{\delta} \boldsymbol{\alpha}\times])^T +R_{b_i}
    \end{split}
\end{equation}
Further decomposition of $Q_{\hat{\lambda}_i}$ in terms of $\boldsymbol{\delta} \boldsymbol{\alpha}$ then yields
\begin{equation}
    Q_{\hat{\lambda}_i}=Q_{\lambda_i}+\delta Q_{\lambda_i}+\delta^2Q_{\lambda_i}\label{eqn:q_lam_i}
\end{equation}
where
\begin{subequations}
    \begin{gather}
    Q_{\lambda_i}=AR_{r_i}A^T-AR_{rb_i}-R_{rb_i}^TA^T+R_{b_i}\\
    \delta Q_{\lambda_i}(\boldsymbol{\delta}\boldsymbol{\alpha})=\mathcal{K}^i[\boldsymbol{\delta}\boldsymbol{\alpha}\times]+[\boldsymbol{\delta}\boldsymbol{\alpha}\times]^T\mathcal{K}^{iT}\\
    \delta^2Q_{\lambda_i}=[\boldsymbol{\delta}\boldsymbol{\alpha}\times]AR_{r_i}A^T[\boldsymbol{\delta}\boldsymbol{\alpha}\times]^T
    \end{gather}
\end{subequations}
with
\begin{equation}
    \mathcal{K}^i=AR_{r_i}A^T-R_{rb_i}^TA^T
\end{equation}
The inverse of $Q_{\hat{\lambda}_i}$ is approximated by
\begin{align}
    \label{iQ_hat}
    Q^{-1}_{\hat{\lambda}_i}=Q^{-1}_{\lambda_i}-Q^{-1}_{\lambda_i}\delta Q_{\lambda_i}Q^{-1}_{\lambda_i}-Q^{-1}_{\lambda_i}\delta^2Q_{\lambda_i}Q^{-1}_{\lambda_i}
\end{align}
Therefore, the matrix $Q^{-1}_{\lambda_i}$ is the only portion of $Q^{-1}_{\hat{\lambda}_i}$ that is not a function of the attitude error.
This portion will be used for building the second-order cost function.
The second summation of the cost function in Eq.~\eqref{eqn:cost_ahat} contains $\Bar{Q}_{\hat{\lambda}}$, which itself is a function of $\boldsymbol{\delta \alpha}$.
The portion of $\Bar{Q}_{\hat{\lambda}}$ that is not dependent on the attitude error needs to be extracted.
This way, the corresponding part of the cost function ignores higher-order terms.
The expansion of $\Bar{Q}_{\hat{\lambda}}$ is given as
\begin{align}
    \Bar{Q}_{\hat{\lambda}}=\Bigg(\sum_{i=1}^{n}Q^{-1}_{\lambda_i}\Bigg)^{-1}+\Bigg(\sum_{i=1}^{n}Q^{-1}_{\lambda_i}\Bigg)^{-1}\Bigg[\sum_{i=1}^n Q^{-1}_{\lambda_i}\bigg(\delta Q^{-1}_{\lambda_i}+\delta^2 Q^{-1}_{\lambda_i}\bigg)Q^{-1}_{\lambda_i}\Bigg]\Bigg(\sum_{i=1}^{n}Q^{-1}_{\lambda_i}\Bigg)^{-1}
\end{align}
The summations in the above equation can be abbreviated as
\begin{subequations}
    \begin{gather}
        S_{\lambda}=\Bigg(\sum_{i=1}^{n}Q^{-1}_{\lambda_i}\Bigg)^{-1}\label{eqn:s_lambda}\\
    \delta S_{\lambda}(\boldsymbol{\delta} \boldsymbol{\alpha})=S_{\lambda}\Bigg(\sum_{i=1}^n Q^{-1}_{\lambda_i}\delta Q^{-1}_{\lambda_i}Q^{-1}_{\lambda_i}\Bigg) S_{\lambda}\\
    \delta^2 S_{\lambda}(\boldsymbol{\delta} \boldsymbol{\alpha}^T\boldsymbol{\delta} \boldsymbol{\alpha})=S_{\lambda}\Bigg(\sum_{i=1}^n Q^{-1}_{\lambda_i}\delta^2 Q^{-1}_{\lambda_i}Q^{-1}_{\lambda_i}\Bigg) S_{\lambda}
    \end{gather}
\end{subequations}
Hence, $S_{\lambda}$ emerges as the only term that is not a function of the attitude error. %Now, we return to the second order approximation of the cost function.
Utilizing the first-order errors in Eq.~\eqref{bti-ati}, the matrix $Q_{\lambda_i}$ in Eq.~\eqref{eqn:q_lam_i}, and $S_{\lambda}$ in Eq.~\eqref{eqn:s_lambda}, the following approximation of the cost function is given:
\begin{equation}
L(\boldsymbol{\delta}\boldsymbol{\alpha},\boldsymbol{\delta}\boldsymbol{\alpha}^T\boldsymbol{\delta}\boldsymbol{\alpha})=\frac{1}{2}\left[L_1(\boldsymbol{\delta}\boldsymbol{\alpha})+L_2(\boldsymbol{\delta}\boldsymbol{\alpha}^T\boldsymbol{\delta}\boldsymbol{\alpha})\right]\label{eqn:def_sec_order_cost}
\end{equation}
where the first- and second-order terms yield
\begin{subequations}
    \begin{gather}
        L_1(\boldsymbol{\delta} \boldsymbol{\alpha})=\boldsymbol{\delta} \boldsymbol{\alpha}^T\mathbf{g}\\
    L_2(\boldsymbol{\delta} \boldsymbol{\alpha}^T\boldsymbol{\delta} \boldsymbol{\alpha})=  \boldsymbol{\delta} \boldsymbol{\alpha}^T \mathcal{H} \boldsymbol{\delta} \boldsymbol{\alpha}
    \end{gather}
\end{subequations}
with
\begin{subequations}
    \begin{gather}
        \mathbf{g}= \left(\sum_{i=1}^n \mathcal{A}_i Q^{-1}_{\lambda_i} \right)S_{\lambda}\left(\sum_{i=1}^n Q^{-1}_{\lambda_i}\boldsymbol{\nu}_i \right)+\sum_{i=1}^n \mathcal{A}^T_i Q^{-1}_{\lambda_i}\boldsymbol{\nu}_i\label{def_G}\\
    \mathcal{H}=\left( \sum_{i=1}^n \mathcal{A}^T_i Q^{-1}_{\lambda_i}\mathcal{A}_i\right)-\left(\sum_{i=1}^n Q^{-1}_{\lambda_i}\mathcal{A}_i \right)^T  S_{\lambda} \left(\sum_{i=1}^n Q^{-1}_{\lambda_i}\mathcal{A}_i\right)\label{H_def}
    \end{gather}
\end{subequations}
% \begin{align}
    
% \end{align}
From the necessary condition for the extremum of the cost function $L$ with respect to $\boldsymbol{\delta} \boldsymbol{\alpha}$, i.e.
\begin{align}
    \frac{\partial L}{\partial \boldsymbol{\delta} \boldsymbol{\alpha}}=\frac{\partial}{\partial \boldsymbol{\delta} \boldsymbol{\alpha}}\big(L_1+L_2\big)=\mathbf{0}
\end{align}
the vector estimate of the attitude-error emanates as
\begin{align}
    \boldsymbol{\delta} \boldsymbol{\alpha}=-\mathcal{H}^{-1}\mathbf{g}
\end{align}
Employing Eq.~\eqref{eqn:def_nu_i}) allows for expanding the above expression to
\begin{equation}
    \boldsymbol{\delta} \boldsymbol{\alpha}=-\mathcal{H}^{-1}\left[\left(\sum_{i=1}^n \mathcal{A}_i Q^{-1}_{\lambda_i} \right)S_{\lambda}\bigg[\sum_{i=1}^n Q^{-1}_{\lambda_i}(\mathbf{p}-\boldsymbol{\Delta} \mathbf{a}_i) \bigg]+\sum_{i=1}^n \mathcal{A}^{T}_i Q^{-1}_{\lambda_i}(\mathbf{p}-\boldsymbol{\Delta} \mathbf{a}_i)\right]
\end{equation}
At the same time, the following relation holds:
\begin{equation}
    \left(\sum_{i=1}^n \mathcal{A}_i Q^{-1}_{\lambda_i} \right)S_{\lambda}\left(\sum_{i=1}^n Q^{-1}_{\lambda_i}\mathbf{p} \right)+\sum_{i=1}^n \mathcal{A}^{T}_i Q^{-1}_{\lambda_i}\mathbf{p}=\mathbf{0}
\end{equation}
Given that $\mathcal{A}^T_i=-\mathcal{A}_i$, as well as that $Q^{-1}_{\lambda_i}$, $S_{\lambda}$ and $\mathcal{H}$ are symmetric, the terms containing the translation vector $\mathbf{p}$ cancel, thus yielding
\begin{align}\label{eqn:d_alpha}
    \boldsymbol{\delta} \boldsymbol{\alpha}=\mathcal{H}^{-1}\left[\left(\sum_{i=1}^n \mathcal{A}_i Q^{-1}_{\lambda_i} \right)S_{\lambda}\left(\sum_{i=1}^n Q^{-1}_{\lambda_i}\boldsymbol{\Delta} \mathbf{a}_i \right)-\sum_{i=1}^n \mathcal{A}_i Q^{-1}_{\lambda_i}\boldsymbol{\Delta} \mathbf{a}_i\right]
\end{align}
This attitude-error now allows for the derivation of the error-covariance expressions.
The estimation error-covariance of the attitude is defined as
\begin{equation}
\begin{split}
    P_{\boldsymbol{\delta} \boldsymbol{\alpha}}&=\mathcal{E}\{\boldsymbol{\delta} \boldsymbol{\alpha}\boldsymbol{\delta} \boldsymbol{\alpha}^T\}\\
    &=\mathcal{E}\Bigg\{\mathcal{H}^{-1}\left[\big(\sum_{i=1}^n \mathcal{A}_i Q^{-1}_{\lambda_i} \big)S_{\lambda}\big(\sum_{i=1}^n Q^{-1}_{\lambda_i}\boldsymbol{\Delta} \mathbf{a}_i \big)-\sum_{i=1}^n \mathcal{A}_i Q^{-1}_{\lambda_i}\boldsymbol{\Delta} \mathbf{a}_i\right]
    \\
    & \times\left[\big(\sum_{i=1}^n \mathcal{A}_i Q^{-1}_{\lambda_i} \big)S_{\lambda}\big(\sum_{i=1}^n Q^{-1}_{\lambda_i}\boldsymbol{\Delta} \mathbf{a}_i \big)-\sum_{i=1}^n \mathcal{A}_i Q^{-1}_{\lambda_i}\boldsymbol{\Delta} \mathbf{a}_i\right]^T\mathcal{H}^{-T}\Bigg\}
\end{split}
\end{equation}
Further expansion of the individual terms leads to
\begin{equation}
\begin{split}
    P_{\boldsymbol{\delta} \boldsymbol{\alpha}}& =\mathcal{H}^{-1}\Bigg[-\sum_{i=1}^n\big(\mathcal{A}_i Q^{-1}_{\lambda_i}\mathcal{E}\{\boldsymbol{\Delta} \mathbf{a}_i\boldsymbol{\Delta} \mathbf{a}^T_i\}Q^{-1}_{\lambda_i}\mathcal{A}_i\big)\\
    &+\big(\sum_{i=1}^n\mathcal{A}_i Q^{-1}_{\lambda_i}\big)S_{\lambda}\big(\sum_{i=1}^nQ^{-1}_{\lambda_i}\mathcal{E}\{\boldsymbol{\Delta} \mathbf{a}_i\boldsymbol{\Delta} \mathbf{a}^T_i\}Q^{-1}_{\lambda_i}\big) S_{\lambda}\big(\sum_{i=1}^n Q^{-1}_{\lambda_i}\mathcal{A}^T_i\big)\\
    &-\big(\sum_{i=1}^n\mathcal{A}_iQ^{-1}_{\lambda_i}\mathcal{E}\{\boldsymbol{\Delta} \mathbf{a}_i\boldsymbol{\Delta} \mathbf{a}^T_i\}Q^{-1}_{\lambda_i}\big)S_{\lambda}\big(\sum_{i=1}^n Q^{-1}_{\lambda_i}\mathcal{A}^T_i\big)\\
    &+\big(\sum_{i=1}^n\mathcal{A}_i Q^{-1}_{\lambda_i}\big)S_{\lambda} \big(\sum_{i=1}^n Q^{-1}_{\lambda_i} \mathcal{E}\{\boldsymbol{\Delta} \mathbf{a}_i\boldsymbol{\Delta} \mathbf{a}^T_i\}Q^{-1}_{\lambda_i}\mathcal{A}_i\big)  \Bigg]\mathcal{H}^{-T}
\end{split}
\end{equation}
Employing the fact that
\begin{subequations}
    \begin{gather}
        P_{\boldsymbol{\Delta} \mathbf{a}_i}\equiv \mathcal{E}\{\boldsymbol{\Delta} \mathbf{a}_i\boldsymbol{\Delta} \mathbf{a}^T_i\}=Q_{\lambda_i}\label{eqn:P_delta_ai}\\
    \mathcal{E}\{\boldsymbol{\Delta} \mathbf{a}_i\boldsymbol{\Delta} \mathbf{a}^T_j\}=0_{3\times3}\ , j\neq i\label{eqn:P_delta_ainot}
    \end{gather}
\end{subequations}
the attitude error-covariance can be rewritten as
\begin{equation}
\begin{split}
    P_{\boldsymbol{\delta} \boldsymbol{\alpha}}&=\mathcal{H}^{-1}\Bigg[-\sum_{i=1}^n\big(\mathcal{A}_iQ^{-1}_{\lambda_i}Q_{\lambda_i}Q^{-1}_{\lambda_i}\mathcal{A}_i\big) \\
    &+\big(\sum_{i=1}^n\mathcal{A}_i Q^{-1}_{\lambda_i}\big)S_{\lambda}\big(\sum_{i=1}^n Q^{-1}_{\lambda_i} Q_{\lambda_i}Q^{-1}_{\lambda_i}\big)S_{\lambda}\big(\sum_{i=1}^n Q^{-1}_{\lambda_i}\mathcal{A}^T_i\big) \\
    &-\big(\sum_{i=1}^n\mathcal{A}_iQ^{-1}_{\lambda_i}Q_{\lambda_i}Q^{-1}_{\lambda_i}\big)S_{\lambda}\big(\sum_{i=1}^n Q^{-1}_{\lambda_i}\mathcal{A}^T_i\big)\\
    &+\big(\sum_{i=1}^n\mathcal{A}_i Q^{-1}_{\lambda_i}\big)S_{\lambda} \big(\sum_{i=1}^n Q^{-1}_{\lambda_i} Q_{\lambda_i}Q^{-1}_{\lambda_i}\mathcal{A}_i\big)  \Bigg]\mathcal{H}^{-T}
\end{split}
\end{equation}
The last two contributions in the above sum cancel out each other, thus yielding
\begin{equation}
    P_{\boldsymbol{\delta} \boldsymbol{\alpha}}=\mathcal{H}^{-1}\Bigg[-\sum_{i=1}^n\big(\mathcal{A}_iQ^{-1}_{\lambda_i}\mathcal{A}_i\big)+\big(\sum_{i=1}^n\mathcal{A}_i Q^{-1}_{\lambda_i}\big)S_{\lambda}\big(\sum_{i=1}^n Q^{-1}_{\lambda_i}\big)S_{\lambda}\big(\sum_{i=1}^n Q^{-1}_{\lambda_i}\mathcal{A}^T_i\big)  \Bigg]\mathcal{H}^{-T}
\end{equation}
With Eqs.~\eqref{eqn:s_lambda} and \eqref{H_def}, the above expression further simplifies to
\begin{equation}
    P_{\boldsymbol{\delta} \boldsymbol{\alpha}}=\mathcal{H}^{-1}\label{eqn:P_delta_alpha_final}
\end{equation}
which verifies that the estimate error-covariance of the attitude error is equal to the Hessian of the cost function. Note that a more detailed discussion of this observation is provided later in the context of the Fisher information matrix (FIM) for the cost function in Eq.~\eqref{eqn:cost_ahat}. 

The estimation error $\boldsymbol{\delta} \mathbf{p}$ for the translation vector is now derived, which begins with 
\begin{equation}
    \hat{\mathbf{p}}=\mathbf{p}+\boldsymbol{\delta} \mathbf{p}\label{phat-p}
\end{equation}
Decomposition of Eq.~\eqref{eqn:def_phat} by utilizing Eq.~\eqref{bti-ati} to separate the first-order terms in the attitude error yields
\begin{equation}
    \mathbf{p}+\boldsymbol{\delta} \mathbf{p}\approx-\left(\sum_{i=1}^{n}Q^{-1}_{\lambda_i}\right)^{-1} \left[\sum_{i=1}^{n}Q^{-1}_{\lambda_i}\left(-\boldsymbol \nu_i-\mathcal{A}_i\boldsymbol{\delta} \boldsymbol{\alpha}\right)\right]
\end{equation}
Thus, the estimate-error $\boldsymbol{\delta} \mathbf{p}$ emerges as
\begin{equation}
    \boldsymbol{\delta} \mathbf{p}=-S_{\lambda}\left(\sum_{i=1}^{n}Q^{-1}_{\lambda_i}\left(-\boldsymbol \nu_i-\mathcal{A}_i\boldsymbol{\delta} \boldsymbol{\alpha}+\mathbf{p}\right)\right)
\end{equation}
From the definition of $\boldsymbol \nu_i$ in Eq.~\eqref{eqn:def_nu_i} gives
\begin{equation}\label{eqn:err_p}
    \boldsymbol{\delta} \mathbf{p}=-S_{\lambda}\sum_{i=1}^{n}Q^{-1}_{\lambda_i}\left(\boldsymbol{\Delta} \mathbf{a}_i-\mathcal{A}_i\boldsymbol{\delta} \boldsymbol{\alpha}\right)
\end{equation}
The error-covariance of the translation vector within the first-order of estimation errors is given by
\begin{equation}
    \begin{split}
        \label{def_cov_phat}
        \text{cov}\{\hat{\mathbf{p}}\}&\equiv \mathcal{E}\left\{(\hat{\mathbf{p}}-\mathcal{E}\{\hat{\mathbf{p}}\})(\hat{\mathbf{p}}-\mathcal{E}\{\hat{\mathbf{p}}\})^T\right\}\\
        &=\mathcal{E}\{\boldsymbol{\delta} \mathbf{p}\boldsymbol{\delta} \mathbf{p}^T\}
    \end{split}
\end{equation}
in which the fact that the estimation for the translation vector is unbiased within the first-order terms of error is used, and $\mathcal{E}\{\hat{\mathbf{p}}\}=\mathbf{p}$. 
Then, the transnational error-covariance is computed as
\begin{equation}
    \text{cov}\{\hat{\mathbf{p}}\}=\mathcal{E}\left\{\left[-S_{\lambda}\sum_{i=1}^{n}Q^{-1}_{\lambda_i}\left(\boldsymbol{\Delta}\mathbf{a}_i-\mathcal{A}_i\boldsymbol{\delta} \boldsymbol{\alpha}\right)\right]\left[-S_{\lambda}\sum_{i=1}^{n}Q^{-1}_{\lambda_i}\left(\boldsymbol{\Delta}\mathbf{a}_i-\mathcal{A}_i\boldsymbol{\delta} \boldsymbol{\alpha}\right)\right]^T\right\}
\end{equation}
It is observed that the cross-covariance of the attitude errors and $\boldsymbol{\Delta} \mathbf{a}_i$ is required for the translation error-covariance, which is computed as
\begin{equation}
    \mathcal{E}\{\boldsymbol{\delta}\boldsymbol{\alpha}\Delta\mathbf{a}^T_i\}=\mathcal{H}^{-1}(\mathcal{A}_i-\Bar{\mathcal{A}})\label{eqn:p_alpha_ai}
\end{equation}
where 
\begin{align}
    \Bar{\mathcal{A}}=S_{\lambda}\sum_{i=1}^n Q^{-1}_{\lambda_i} \mathcal{A}_i
\end{align}
Using Eqs.~\eqref{eqn:P_delta_ai}, \eqref{eqn:P_delta_ainot} and \eqref{eqn:p_alpha_ai}, and from the attitude estimate error-covariance in Eq.~\eqref{eqn:P_delta_alpha_final}, the translation vector error-covariance is now given by
\begin{align}
    \text{cov}\{\hat{\mathbf{p}}\}=S_{\lambda}+\Bar{\mathcal{A}}P_{\boldsymbol{\delta}\alpha}\Bar{\mathcal{A}}^T\label{calc_cov_phat}
\end{align}

Because there are estimates for the observation vectors from the TLS formulation in Eq.~\eqref{eqn:vecdhat}, their associated covariance expressions can be derived. First an expression for their corresponding first-order residuals is found, and then this residual approximation is used to construct an analytical covariance formulation. Using Eq.~\eqref{eqn:vecdhat} for the observation vectors and using the derivation in \cite{cheng2021optimal}, it can be shown that the estimates of the observation vectors are
\begin{subequations}
    \begin{gather}
        \hat{\mathbf{b}}_i=\tilde{\mathbf{b}}_i+(R_{rb_i}^T\hat{A}^T-R_{b_i})Q^{-1}_{\hat{\lambda_i}}(\tilde{\mathbf{b}}_i-\hat{A}\tilde{\mathbf{r}}_i+\hat{\mathbf{p}}) \label{eqn:bhat-btilde}\\
    \hat{\mathbf{r}}_i=\tilde{\mathbf{r}}_i+(R_{r_i}\hat{A}^T-R_{rb_i})Q^{-1}_{\hat{\lambda_i}}(\tilde{\mathbf{b}}_i-\hat{A}\tilde{\mathbf{r}}_i+\hat{\mathbf{p}}) \label{eqn:rhat-rtilde}
    \end{gather}
\end{subequations}
Define the estimate-error for the observation vectors:
\begin{subequations}
    \begin{gather}
        \boldsymbol{\delta}\mathbf{b}_i=\hat{\mathbf{b}}_i-\mathbf{b}_i\\
    \boldsymbol{\delta}\mathbf{r}_i=\hat{\mathbf{r}}_i-\mathbf{r}_i
    \end{gather}
\end{subequations}
The residual errors using Eq.~\eqref{eqn:r_tilde_def} and \eqref{eqn:b_tilde_def} are $\boldsymbol{\delta}\mathbf{b}_i-\boldsymbol{\Delta}\mathbf{b}_i$ and $\boldsymbol{\delta}\mathbf{r}_i-\boldsymbol{\Delta}\mathbf{r}_i$. Then deducting both sides of Eq.~\eqref{eqn:bhat-btilde} by $\mathbf{b}_i$ and Eq.~\eqref{eqn:rhat-rtilde} by $\mathbf{r}_i$ leads to
\begin{subequations}
    \begin{gather}
       \boldsymbol{\delta}\mathbf{b}_i-\boldsymbol{\Delta}\mathbf{b}_i=(R_{rb_i}^T\hat{A}^T-R_{b_i})Q^{-1}_{\hat{\lambda_i}}(\tilde{\mathbf{b}}_i-\hat{A}\tilde{\mathbf{r}}_i+\hat{\mathbf{p}})\\
    \boldsymbol{\delta}\mathbf{r}_i-\boldsymbol{\Delta}\mathbf{r}_i=(R_{r_i}\hat{A}^T-R_{rb_i})Q^{-1}_{\hat{\lambda_i}}(\tilde{\mathbf{b}}_i-\hat{A}\tilde{\mathbf{r}}_i+\hat{\mathbf{p}}) 
    \end{gather}
\end{subequations}
Their corresponding first-order approximations are given by
\begin{subequations}
    \begin{gather}
        \boldsymbol{\delta}\mathbf{b}_i-\boldsymbol{\Delta}\mathbf{b}_i\approx C_i(\boldsymbol{\Delta}\mathbf{a}_i-\mathcal{G}_i\boldsymbol{\delta}\mathbf{f})\label{apprxb-btilde}\\
    \boldsymbol{\delta}\mathbf{r}_i-\boldsymbol{\Delta}\mathbf{r}_i\approx D_i(\boldsymbol{\Delta}\mathbf{a}_i-\mathcal{G}_i\boldsymbol{\delta}\mathbf{f})\label{apprxr-rtilde}
    \end{gather}
\end{subequations}
where
\begin{subequations}
    \begin{align}
        C_i&=(R_{rb_i}^T A^T-R_{b_i})Q^{-1}_{\lambda_i}\\
    D_i&=(R_{r_i} A^T-R_{rb_i})Q^{-1}_{\lambda_i}\\
    \mathcal{G}_i&=\begin{bmatrix}\mathcal{A}_i &  -I_{3\times3}\end{bmatrix}\label{eqn:def_Gi}\\
    \boldsymbol{\delta}\mathbf{f}&=\begin{bmatrix}\boldsymbol{\delta}\boldsymbol{\alpha} \\ \boldsymbol{\delta} \mathbf{p}\end{bmatrix}_{6\times 1}\label{eqn:def_delta_f}
    \end{align}
\end{subequations}
Using
\begin{subequations}
    \begin{gather}
        P_{\boldsymbol{\Delta} \mathbf{a}_i}=Q_{\lambda_i}\\
        P_{\boldsymbol{\Delta}\mathbf{a}_i \boldsymbol{\Delta}\Bar{\mathbf{a}}}=S_{\lambda}\\
        P_{ \boldsymbol{\Delta}\Bar{\mathbf{a}} \boldsymbol{\delta}\boldsymbol{\alpha}} =0_{3\times 3}
    \end{gather}
\end{subequations}
 and from Eq.~\eqref{eqn:p_alpha_ai}, the covariances of the measurement residuals are given by
 \begin{subequations}
     \begin{align}
         \text{cov}(\hat{\mathbf{b}}_i-\tilde{\mathbf{b}}_i)&=\mathcal{E}\{(\boldsymbol{\delta}\mathbf{b}_i-\boldsymbol{\Delta}\mathbf{b}_i)(\boldsymbol{\delta}\mathbf{b}_i-\boldsymbol{\Delta}\mathbf{b}_i)^T\}\nonumber\\
    &=C_i (Q_{\lambda_i}-\mathcal{G}_i P_{\mathbf{f}} \mathcal{G}_i^T) C^T_i\label{eqn:cov_bi}\\
    \text{cov}(\hat{\mathbf{r}}_i-\tilde{\mathbf{r}}_i)&=\mathcal{E}\{(\boldsymbol{\delta}\mathbf{r}_i-\boldsymbol{\Delta}\mathbf{r}_i)(\boldsymbol{\delta}\mathbf{r}_i-\boldsymbol{\Delta}\mathbf{r}_i)^T\}\nonumber\\
    &=D_i (Q_{\lambda_i}-\mathcal{G}_i P_{\mathbf{f}} \mathcal{G}_i^T) D^T_i\label{eqn:cov_ri}
     \end{align}
 \end{subequations}
where
\begin{equation}
        P_{\mathbf{f}}\equiv\text{cov}\{\boldsymbol{\delta}\mathbf{f}\}=\left(\sum_{i=1}^n \mathcal{G}_i^T Q^{-1}_{\lambda_i} \mathcal{G}_i\right)^{-1}
\end{equation}
Equations\eqref{apprxb-btilde} and \eqref{apprxr-rtilde} can now be written as
\begin{subequations}
    \begin{gather}
        \boldsymbol{\delta}\mathbf{b}_i\approx\boldsymbol{\Delta}\mathbf{b}_i+ C_i(\boldsymbol{\Delta}\mathbf{a}_i-\mathcal{G}_i\boldsymbol{\delta}\mathbf{f})\\
    \boldsymbol{\delta}\mathbf{r}_i\approx\boldsymbol{\Delta}\mathbf{r}_i+ D_i(\boldsymbol{\Delta}\mathbf{a}_i-\mathcal{G}_i\boldsymbol{\delta}\mathbf{f})
    \end{gather}
\end{subequations}
Their corresponding estimate covariances are given by
\begin{equation}
\begin{split}
    P_{b_i}&=\mathcal{E}\{\boldsymbol{\delta}\mathbf{b}_i\boldsymbol{\delta}\mathbf{b}_i^T\}\\
    &=\mathcal{E}\{(\hat{\mathbf{b}}_i-\mathbf{{b}}_i)(\hat{\mathbf{b}}_i-\mathbf{{b}}_i)^T\}\\
    &=R_{b_i}+\text{cov}\{\hat{\mathbf{b}}_i-\tilde{\mathbf{b}}_i\}\\
    &+C_iR_{b_i}-C_i\mathcal{G}_i P_{\mathbf{f}}\mathcal{G}_i^T Q^{-1}_{\lambda_i} R_{b_i}\\
    &+\left(C_iR_{b_i}-C_i\mathcal{G}_i P_{\mathbf{f}} \mathcal{G}_i^T Q^{-1}_{\lambda_i} R_{b_i}\right)^T\label{eqn:P_bi}
\end{split}
\end{equation}
\begin{equation}
    \begin{split}
        P_{r_i}&=\mathcal{E}\{\boldsymbol{\delta}\mathbf{r}_i\delta\mathbf{r}^T_i\}\\
    &=\mathcal{E}\{(\hat{\mathbf{r}}_i-\mathbf{{r}}_i)(\hat{\mathbf{r}}_i-\mathbf{{r}}_i)^T\}\\
    &=R_{r_i}+\text{cov}\{\hat{\mathbf{r}}_i-\tilde{\mathbf{r}}_i\}\\
    &-D_iAR_{r_i}+D_i\mathcal{G}_i P_{\mathbf{f}}\mathcal{G}_i^T Q^{-1}_{\lambda_i} AR_{r_i}\\
    &+\left(-D_iAR_{r_i}+D_i\mathcal{G}_i P_{\mathbf{f}}\mathcal{G}_i^T Q^{-1}_{\lambda_i} AR_{r_i}\right)^T\label{eqn:P_ri}
    \end{split}
\end{equation}

% in which 
% \begin{equation}
%         \text{cov}\{\boldsymbol{\delta}\mathbf{x}\}\equiv E\{\boldsymbol{\delta}\mathbf{x}\boldsymbol{\delta}\mathbf{x}^T\}=\left(\sum_{i=1}^n \mathcal{G}_i Q^{-1}_{\lambda_i} \mathcal{G}^T_i\right)^{-1}
% \end{equation}
\subsection{Isotropic Noise Covariance}
The previous section provided analytical expressions for the covariance of estimates and residuals for the most generic case of observation noise covariance, which is a fully-populated symmetric positive-definite matrix as denoted in Eq.~\eqref{eqn:R_i}
Now a particular case of previous covariance derivations is elaborated.
In some sensors, there is a simplifying assumption on the distribution of noise having the same characteristics in the different space coordinates of $x$, $y$ and $z$.
This presumption results in a particular form of the noise covariance called \textit{isotropic}.
For the pose estimation problem in this work, this isotropic covariance of observation errors is denoted by
\begin{equation}
R_i=\begin{bmatrix}\sigma^2_{r_i}I_{3\times3} & 0_{3\times3}\\
    0_{3\times3} & \sigma^2_{b_i}I_{3\times3}\end{bmatrix}
\end{equation}
in which $\sigma_{r_i}$ and $\sigma_{b_i}$ indicate the standard deviation of noise for the sensors in the reference and body frames, respectively.
The cross-correlation between the reference and body sensor noise will be zero, as shown in the off-diagonal blocks.
For the isotropic case, the weight matrix $Q_{\hat{\lambda}_i}$ in the cost function from Eq.~\eqref{eqn:expand_Q} shrinks to
\begin{equation}
\begin{split}
    Q_{\hat{\lambda}_i}&=\hat{A}R_{r_i}\hat{A}^T+R_{b_i}\\
    &=\hat{A}\sigma^2_{r_i}I_{3\times3}\hat{A}^T+\sigma^2_{b_i}I_{3\times3}\\
    &=\left(\sigma^2_{r_i}+\sigma^2_{b_i}\right)I_{3\times3}\\
    &=\sigma^2_i I_{3\times3}
\end{split}
\end{equation}
where
\begin{equation}
    \sigma_i=\sqrt{\sigma^2_{r_i}+\sigma^2_{b_i}}
\end{equation}
As a result, the second weight $\bar{Q}_{\hat{\lambda}}$ of the cost function in Eq.~\eqref{q_cap_bar} becomes
\begin{subequations}
    \begin{align}
        \bar{Q}_{\hat{\lambda}}&=\bar{\sigma}^{-2}I_{3\times3}\\
        \bar{\sigma}&\equiv\sqrt{\sum_{i=1}^n \sigma^{-2}_i}=\sqrt{\sum_{i=1}^n \frac{1}{\sigma^2_{r_i}+\sigma^2_{b_i}} }
    \end{align}
\end{subequations}
Then the attitude-only cost function in Eq.~\eqref{eqn:cost_ahat} yields
\begin{subequations}
\begin{gather}
    J(\hat{A})=\frac{1}{2}\Bigg[\sum_{i=1}^{n}\sigma^{-2}_i\|\tilde{\mathbf{b}}_i-\hat{A}\tilde{\mathbf{r}}_i\|^2\Bigg]-\frac{1}{2}\bar{\sigma}^2\|\Bar{\tilde{\mathbf{b}}}-\hat{A}\Bar{\tilde{\mathbf{r}}}\|^2\label{eqn:cost_iso}\\
    \Bar{\tilde{\mathbf{b}}}=\bar{\sigma}^{-2}\sum_{i=1}^n \sigma^{-2}_i \tilde{\mathbf{b}}_i\\
    \Bar{\tilde{\mathbf{r}}}=\bar{\sigma}^{-2}\sum_{i=1}^n \sigma^{-2}_i \tilde{\mathbf{r}}_i
\end{gather}
\end{subequations}
It is to be noted that in the isotropic case the matrix $Q_{\hat{\lambda}_i}$ is not a function of attitude matrix. Therefore, the attitude error avoids several complications in the derivation.
This simplicity comes with the cost of ignoring the possible cross-covariances in the reference and body frames and assuming the same statistical properties of noise along all space coordinates.
The cost function in Eq.~\eqref{eqn:cost_iso} is already second-order in terms of the attitude matrix and there is no need for trimming the higher-order terms, which is another simplicity resulting from the isotropic assumption.

The optimal attitude from Eq.~\eqref{eqn:d_alpha} yields
\begin{equation}
\begin{split}
    \boldsymbol{\delta \alpha}&=\mathcal{H}^{-1}\left[\sum_{j=1}^n \sigma^{-2}_j\mathcal{A}_j\bar{\sigma}^{-2} \sum_{i=1}^n \sigma^{-2}_i\boldsymbol{\Delta}\mathbf{a}_i-\sum_{i=1}^n\sigma^{-2}_i\mathcal{A}_i\boldsymbol{\Delta}\mathbf{a}_i\right]\\
    &=\mathcal{H}^{-1}\left[\sum_{i=1}^n \sigma^{-2}_i\bar{\mathcal{A}}\boldsymbol{\Delta}\mathbf{a}_i-\sum_{i=1}^n\sigma^{-2}_i\mathcal{A}_i\boldsymbol{\Delta}\mathbf{a}_i\right]\\
    &=\mathcal{H}^{-1}\left[\sum_{i=1}^n \sigma^{-2}_i\left(\bar{\mathcal{A}}-\mathcal{A}_i\right)\boldsymbol{\Delta}\mathbf{a}_i\right]
\end{split}
\end{equation}
in which
\begin{subequations}
\begin{gather}
    \mathcal{A}_i=[A\mathbf{r}_i\times]\\
    \bar{\mathcal{A}}=\bar{\sigma}^{-2}\sum_{i=1}^n \sigma^{-2}_i[A\mathbf{r}_i\times]
\end{gather}
\end{subequations}
and the Hessian matrix $\mathcal{H}$ will be
\begin{equation}\label{Hess_iso}
    \begin{split}
        \mathcal{H}&=\sum_{i=1}^n\sigma^{-2}_i\mathcal{A}^T_i\mathcal{A}_i-\left(\sum_{i=1}^n\sigma^{-2}_i\mathcal{A}_i\right)^T\bar{\sigma}^{-2}\left(\sum_{i=1}^n\sigma^{-2}_i\mathcal{A}_i\right)\\
        &=-\sum_{i=1}^n\sigma^{-2}_i\mathcal{A}^2_i-\left(\bar{\sigma}^{-2}\sum_{i=1}^n\sigma^{-2}_i\mathcal{A}_i\right)^T\bar{\sigma}^{2}\left(\bar{\sigma}^{-2}\sum_{i=1}^n\sigma^{-2}_i\mathcal{A}_i\right)\\
        &=-\sum_{i=1}^n\sigma^{-2}_i\mathcal{A}^2_i+\bar{\sigma}^{2}\bar{\mathcal{A}}^2
    \end{split}
\end{equation}
The skew-symmetric property of matrices $\mathcal{A}_i$ and $\bar{\mathcal{A}}$ is employed in the above derivation, which originates from the skew-symmetric nature of the cross product matrices.

Now the attitude error-covariance can be computed.
From Eq.~\eqref{eqn:P_delta_alpha_final} and Eq.~\eqref{Hess_iso}, this becomes
\begin{equation}
    P_{\boldsymbol{\delta\alpha}}=\left[-\sum_{i=1}^n\sigma^{-2}_i\mathcal{A}^2_i+\bar{\sigma}^{2}\bar{\mathcal{A}}^2\right]^{-1}
\end{equation}
Regarding the position estimates, the estimate errors are derived from Eq.~\eqref{eqn:err_p} as
\begin{equation}
    \begin{split}
        \boldsymbol{\delta} \mathbf{p}&=-\bar{\sigma}^{-2}\sum_{i=1}^{n}\sigma^{-2}_i\left(\boldsymbol{\Delta} \mathbf{a}_i-\mathcal{A}_i\boldsymbol{\delta} \boldsymbol{\alpha}\right) \\
    &=-\boldsymbol{\Delta}\bar{\mathbf{a}}+\bar{\mathcal{A}}\boldsymbol{\delta\alpha}\\
    \end{split}
\end{equation}
where
\begin{equation}
    \begin{split}
      \boldsymbol{\Delta}\bar{\mathbf{a}}&\equiv\bar{\sigma}^{-2}\sum_{i=1}^n\sigma^{-2}_i\boldsymbol{\Delta}\mathbf{a}_i
    \end{split}
\end{equation}    
The resulting error-covariance is obtained from Eq.~\eqref{calc_cov_phat}.
For the isotropic case, it will be
\begin{equation}
    \text{cov}\{\hat{\mathbf{p}}\}=\bar{\sigma}^{-2}I_{3\times3}-\bar{\mathcal{A}}\left[-\sum_{i=1}^n\sigma^{-2}_i\mathcal{A}^2_i+\bar{\sigma}^{2}\bar{\mathcal{A}}^2\right]^{-1}\bar{\mathcal{A}}
\end{equation}
The observation vector estimates can be computed from Eq.~\eqref{eqn:bhat-btilde} and Eq.~\eqref{eqn:rhat-rtilde}, and for the isotropic covariance will result in
\begin{subequations}
\begin{gather}
    \hat{\mathbf{b}}_i=\tilde{\mathbf{b}}_i-\frac{\sigma^2_{b_i}}{\sigma^2_{b_i}+\sigma^2_{r_i}}(\tilde{\mathbf{b}}_i-\hat{A}\tilde{\mathbf{r}}_i+\hat{\mathbf{p}})\\
    \hat{\mathbf{r}}_i=\tilde{\mathbf{r}}_i+\frac{\sigma^2_{r_i}}{\sigma^2_{r_i}+\sigma^2_{b_i}}\hat{A}^T(\tilde{\mathbf{b}}_i-\hat{A}\tilde{\mathbf{r}}_i+\hat{\mathbf{p}})
\end{gather}
\end{subequations}
From the first-order approximation of observation residuals in Eqs.~\eqref{apprxb-btilde} and \eqref{apprxr-rtilde}, the observation residuals are given as
\begin{subequations}
\begin{gather}
    \boldsymbol{\delta}\mathbf{b}_i-\boldsymbol{\Delta}\mathbf{b}_i\approx -\frac{\sigma^2_{b_i}}{\sigma^2_{b_i}+\sigma^2_{r_i}}(\boldsymbol{\Delta}\mathbf{a}_i-\mathcal{G}_i\boldsymbol{\delta}\mathbf{f})\\
    \boldsymbol{\delta}\mathbf{r}_i-\boldsymbol{\Delta}\mathbf{r}_i\approx \frac{\sigma^2_{r_i}}{\sigma^2_{b_i}+\sigma^2_{r_i}}A^T(\boldsymbol{\Delta}\mathbf{a}_i-\mathcal{G}_i\boldsymbol{\delta}\mathbf{f})
\end{gather}
\end{subequations}
with the definitions of $\boldsymbol{\Delta}\mathbf{a}_i$, $\mathcal{G}_i$, and $\boldsymbol{\delta}\mathbf{f}$ in Eqs.~\eqref{eqn:def_Dai}, \eqref{eqn:def_Gi} and \eqref{eqn:def_delta_f}, respectively.
Then, the observation residual covariances become
\begin{subequations}
\begin{align}
    \text{cov}(\hat{\mathbf{b}}_i-\tilde{\mathbf{b}}_i)&=\frac{\sigma^2_{b_i}}{\left(\sigma^2_{b_i}+\sigma^2_{r_i}\right)^2} \left[\left(\sigma^2_{b_i}+\sigma^2_{r_i}\right)I_{3\times3}-\begin{bmatrix}\mathcal{A}_i&-I_{3\times3}\end{bmatrix} \left(\sum_{i=1}^n\frac{1}{\sigma^2_{b_i}+\sigma^2_{r_i}}\begin{bmatrix}-\mathcal{A}^2_i&\mathcal{A}_i\\\mathcal{A}^T_i&I_{3\times3}\end{bmatrix}\right)^{-1} \begin{bmatrix}\mathcal{A}^T_i\\-I_{3\times3}\end{bmatrix}\right]\\
    \text{cov}(\hat{\mathbf{r}}_i-\tilde{\mathbf{r}}_i)&=\frac{\sigma^2_{r_i}}{\left(\sigma^2_{b_i}+\sigma^2_{r_i}\right)^2}A^T \left[\left(\sigma^2_{b_i}+\sigma^2_{r_i}\right)I_{3\times3}-\begin{bmatrix}\mathcal{A}_i&-I_{3\times3}\end{bmatrix} \left(\sum_{i=1}^n\frac{1}{\sigma^2_{b_i}+\sigma^2_{r_i}}\begin{bmatrix}-\mathcal{A}^2_i&\mathcal{A}_i\\\mathcal{A}^T_i&I_{3\times3}\end{bmatrix}\right)^{-1} \begin{bmatrix}\mathcal{A}^T_i\\-I_{3\times3}\end{bmatrix}\right] A
\end{align}
\end{subequations}
The estimate covariances of the observation vectors are simplified from Eqs.~\eqref{eqn:P_bi} and \eqref{eqn:P_ri} as
\begin{equation}
    \begin{split}
        P_{b_i}&=\sigma^2_{b_i}I_{3\times3}+\text{cov}\{\hat{\mathbf{b}}_i-\tilde{\mathbf{b}}_i\}\\
    &-2\frac{\sigma^3_{b_i}}{\sigma^2_{b_i}+\sigma^2_{r_i}}I_{3\times3}+\frac{\sigma^3_{b_i}}{\left(\sigma^2_{b_i}+\sigma^2_{r_i}\right)^2}\begin{bmatrix}\mathcal{A}_i&-I_{3\times3}\end{bmatrix} \left(\sum_{i=1}^n\frac{1}{\sigma^2_{b_i}+\sigma^2_{r_i}}\begin{bmatrix}-\mathcal{A}^2_i&\mathcal{A}_i\\\mathcal{A}^T_i&I_{3\times3}\end{bmatrix}\right)^{-1} \begin{bmatrix}\mathcal{A}^T_i\\-I_{3\times3}\end{bmatrix}\\
    &+\frac{\sigma^3_{b_i}}{\left(\sigma^2_{b_i}+\sigma^2_{r_i}\right)^2}\begin{bmatrix}\mathcal{A}_i&-I_{3\times3}\end{bmatrix} \left(\sum_{i=1}^n\frac{1}{\sigma^2_{b_i}+\sigma^2_{r_i}}\begin{bmatrix}-\mathcal{A}^2_i&\mathcal{A}_i\\\mathcal{A}^T_i&I_{3\times3}\end{bmatrix}\right)^{-T} \begin{bmatrix}\mathcal{A}^T_i\\-I_{3\times3}\end{bmatrix}
    \end{split}
\end{equation}
\begin{equation}
    \begin{split}
        P_{r_i}&=\sigma^2_{r_i}I_{3\times3}+\text{cov}\{\hat{\mathbf{r}}_i-\tilde{\mathbf{r}}_i\} \\
    &-2\frac{\sigma^3_{r_i}}{\sigma^2_{b_i}+\sigma^2_{r_i}}I_{3\times3}+\frac{\sigma^3_{r_i}}{\left(\sigma^2_{b_i}+\sigma^2_{r_i}\right)^2}A^T\begin{bmatrix}\mathcal{A}_i&-I_{3\times3}\end{bmatrix} \left(\sum_{i=1}^n\frac{1}{\sigma^2_{b_i}+\sigma^2_{r_i}}\begin{bmatrix}-\mathcal{A}^2_i&\mathcal{A}_i\\\mathcal{A}^T_i&I_{3\times3}\end{bmatrix}\right)^{-1} \begin{bmatrix}\mathcal{A}^T_i\\-I_{3\times3}\end{bmatrix}A\\
    &+\frac{\sigma^3_{b_i}}{\left(\sigma^2_{b_i}+\sigma^2_{r_i}\right)^2}A^T\begin{bmatrix}\mathcal{A}_i&-I_{3\times3}\end{bmatrix} \left(\sum_{i=1}^n\frac{1}{\sigma^2_{b_i}+\sigma^2_{r_i}}\begin{bmatrix}-\mathcal{A}^2_i&\mathcal{A}_i\\\mathcal{A}^T_i&I_{3\times3}\end{bmatrix}\right)^{-T} \begin{bmatrix}\mathcal{A}^T_i\\-I_{3\times3}\end{bmatrix}A
    \end{split}
\end{equation}
All of the covariance expressions reduce down to the ones derived in \cite{cheng2021optimal}.

\subsection{Fisher Information Matrix}
From the analysis in section \ref{cov_analysis}, the estimate error-covariances for the attitude error $\boldsymbol{\delta} \boldsymbol{\alpha}$, the translation vector $\mathbf{p}$, the residuals as well as the estimate covariances for the observation vectors $\tilde{\mathbf{b}}_i$ and $\tilde{\mathbf{r}}_i$ have been derived. An efficiency proof for the estimate error-covariances of the attitude and translation vector is now shown based on the FIM and the CRLB \cite{cramer1999mathematical} defined for the estimation covariances. For an unbiased estimator $\hat{\mathbf{f}}$, the estimate error-covariance has a lower bound as
\begin{equation}
    \text{cov}\{\hat{\mathbf{f}}\}\geq F^{-1}\equiv \left[\mathcal{E}\left\{-\frac{\partial^2}{\partial \hat{\mathbf{f}} \partial \hat{\mathbf{f}}^T}p(\mathbf{\tilde{y}|\hat{\mathbf{f}}})\right\}\right]^{-1}\label{eqn:crlb}
\end{equation}
The term inside of the expectation shows the Hessian of the the negative log-likelihood function, which is given in Eq.~\eqref{eqn:def_sec_order_cost}.
For an optimal estimator, the equality in the Eq.~\eqref{eqn:crlb} is given, and the estimator is {\it efficient}. The Hessian of the second-order approximated cost function $L$ is the FIM. From Eqs.~\eqref{bti-ati} and Eq.~\eqref{phat-p}, the following are given:
\begin{equation}
\begin{split}
    \tilde{\mathbf{b}}_i-\hat{A}\tilde{\mathbf{r}}_i+\hat{\mathbf{p}}&\approx-\mathbf{p}+\boldsymbol{\Delta}\mathbf{a}_i-\mathcal{A}_i\boldsymbol{\delta}\boldsymbol{\alpha}+\mathbf{p}+\boldsymbol{\delta} \mathbf{p}\\
    &=\boldsymbol{\Delta} \mathbf{a}_i-\mathcal{A}_i\boldsymbol{\delta}\boldsymbol{\alpha}+\boldsymbol{\delta} \mathbf{p} \label{del_brp}
\end{split}
\end{equation}
The second-order cost function involving $\boldsymbol{\delta}{\mathbf{p}}$ is now given by
\begin{equation}
    L =\frac{1}{2}\sum_{i=1}^n\left(\boldsymbol{\Delta} \mathbf{a}_i-\mathcal{A}_i\boldsymbol{\delta}\boldsymbol{\alpha}+\boldsymbol{\delta} \mathbf{p}\right)^T Q^{-1}_{\lambda_i}\left(\boldsymbol{\Delta} \mathbf{a}_i-\mathcal{A}_i\boldsymbol{\delta}\boldsymbol{\alpha}+\boldsymbol{\delta} \mathbf{p}\right)
\end{equation}
The FIM, denoted by $F$, will be
\begin{align}
    F=\begin{bmatrix} \sum_{i=1}^{n}{\mathcal{A}_i^T Q_{\lambda_i}^{-1}\mathcal{A}_i} & -\sum_{i=1}^{n}{\mathcal{A}_i^T Q_{\lambda_i}^{-1}}\\
     -\sum_{i=1}^{n}{Q_{\lambda_i}^{-1}\mathcal{A}_i}& \sum_{i=1}^{n}Q_{\lambda_i}^{-1}\end{bmatrix}
\end{align}
The block matrices of the FIM are
\begin{align}
    F=\begin{bmatrix} F_{11} & F_{12}\\
     F_{21}& F_{22}\end{bmatrix}
\end{align}
The inverse of FIM, denoted by $\mathcal{F}$, is given by
\begin{align}
    \mathcal{F}=\begin{bmatrix} \mathcal{F}_{11} & \mathcal{F}_{12}\\
    \mathcal{F}_{21}& \mathcal{F}_{22}\end{bmatrix}
\end{align}
The $\mathcal{F}_{11}$ block of this matrix is calculated by the Sherman–Morrison–Woodbury lemma \cite{sherman1950adjustment}:
\begin{equation}
\begin{split}
    \mathcal{F}_{11}&=\left(F_{11}-F_{12}F^{-1}_{22}F_{21}\right)^{-1}\\
    &=\left[\sum_{i=1}^{n}{\mathcal{A}_i^T Q_{\lambda_i}^{-1}\mathcal{A}_i}-\left(-\sum_{i=1}^{n}{\mathcal{A}_i^T Q_{\lambda_i}^{-1}}\right)S_{\lambda}\left(-\sum_{i=1}^{n}{Q_{\lambda_i}^{-1}\mathcal{A}_i}\right)\right]^{-1}\\
    &=\mathcal{H}^{-1}
\end{split}
\end{equation}
This has already been proven in Eq.~\eqref{eqn:P_delta_alpha_final}, which is the CRLB for the attitude error.
The term $\mathcal{F}_{22}$ is also derived by using matrix inversion lemma as
\begin{equation}
\begin{split}
    \mathcal{F}_{22}&=\left(F_{22}-F_{21}F^{-1}_{11}F_{12}\right)^{-1}\\
    &=\left[\left(\sum_{i=1}^{n}Q_{\lambda_i}^{-1}\right)-\left(-\sum_{i=1}^{n}Q_{\lambda_i}^{-1}\mathcal{A}_i\right)\left(\sum_{i=1}^{n}\mathcal{A}_i^T Q_{\lambda_i}^{-1}\mathcal{A}_i\right)^{-1}
    \left(-\sum_{i=1}^{n}\mathcal{A}_i^T Q_{\lambda_i}^{-1}\right)\right]^{-1}\\
    &=\left[S^{-1}_{\lambda}-S^{-1}_{\lambda}\bar{\mathcal{A}}\left(\sum_{i=1}^{n}\mathcal{A}_i^T Q_{\lambda_i}^{-1}\mathcal{A}_i\right)^{-1}\bar{\mathcal{A}}^TS^{-1}_{\lambda}\right]^{-1}
\end{split}
\end{equation}
Using Eq.~\eqref{eqn:P_delta_alpha_final} leads to
\begin{equation}
\begin{split}
    \mathcal{F}_{22}&=S_{\lambda}+\bar{\mathcal{A}}\left[\sum_{i=1}^{n}\mathcal{A}_i^T Q_{\lambda_i}^{-1}\mathcal{A}_i-\sum_{i=1}^{n}\mathcal{A}_i^T Q_{\lambda_i}^{-1} S_{\lambda}\sum_{i=1}^{n} Q_{\lambda_i}^{-1}\mathcal{A}_i\right]^{-1}\bar{\mathcal{A}}^T\\
    &=S_{\lambda}+\bar{\mathcal{A}}\text{cov}\{\boldsymbol{\delta}\boldsymbol{\alpha}\}\bar{\mathcal{A}}^T
\end{split}
\end{equation}
This proves the CRLB  for covariance of $\{\boldsymbol{\delta} \mathbf{p}\}$. Note that from Eq.~\eqref{eqn:P_delta_alpha_final} it can be concluded that the CRLB holds for the attitude estimate because the estimate error-covariance is equal to the inverse of the Hessian of the cost function in Eq.~\eqref{eqn:def_sec_order_cost}.

% \section{Algorithm for Estimation and Covariance Analysis}\label{pseudo_code}

% \begin{figure}
%   \hspace*{-1.5cm}
%   \centering
%   \includegraphics[width=7.5in]{bes.eps}\\
%   \caption{Monte Carlo simulation for for estimates and residuals of the vector observation $\mathbf{b_1}$.}\label{fig:bes}
%   \hspace{1cm}
%   \centering
%   \hspace*{-1.5cm}
%   \includegraphics[width=7.5in]{res.eps}\\
%   \caption{Monte Carlo simulation for estimates and residuals of the vector observation $\mathbf{r_1}$.}\label{fig:res}
% \end{figure}

\section{Numerical Validation}\label{sec_monte_carlo}
\begin{figure}[h!]
  \begin{centering}
      \subfigure[{\bf Attitude Errors}]
      {\includegraphics[width=0.49\textwidth]{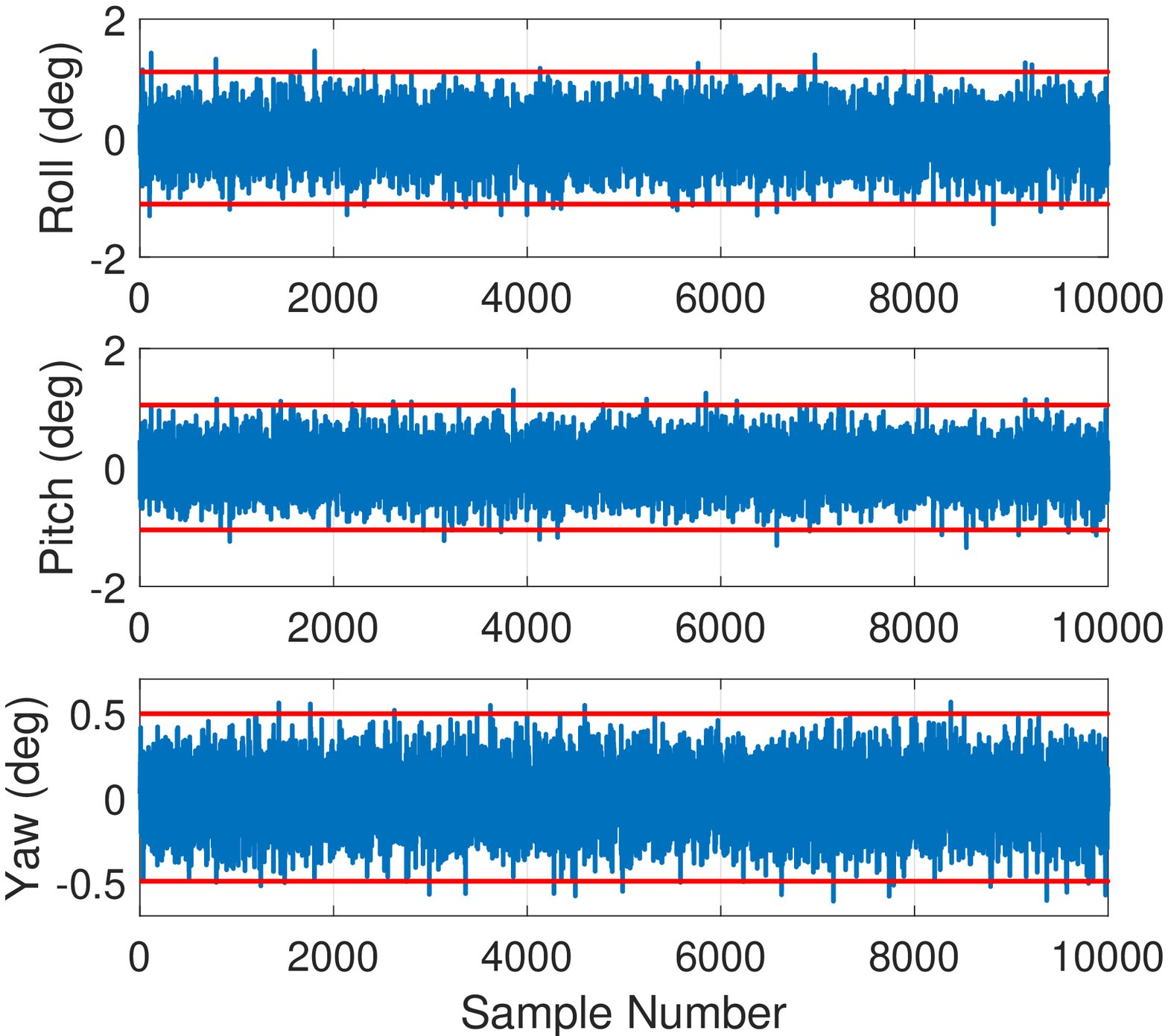}\label{fig:rpy}}
            \subfigure[{\bf Position Errors}]
      {\includegraphics[width=0.49\textwidth]{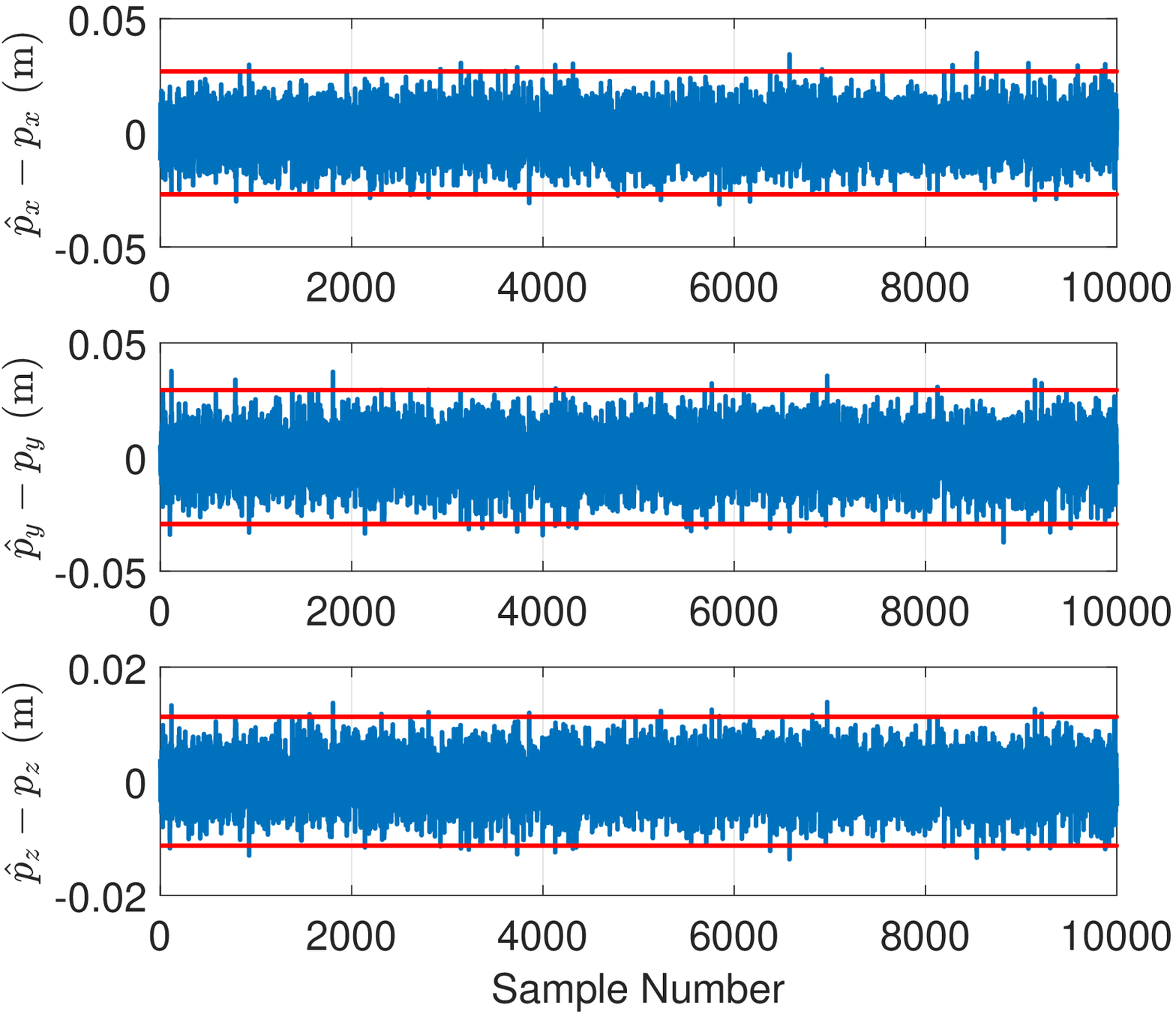}\label{fig:p}}
    \caption{Monte Carlo simulation for the attitude and translation vector.}
  \end{centering}
\end{figure}

\begin{figure}[h!]
  \begin{centering}
      \subfigure[{\bf Estimation Errors}]
      {\includegraphics[width=0.49\textwidth]{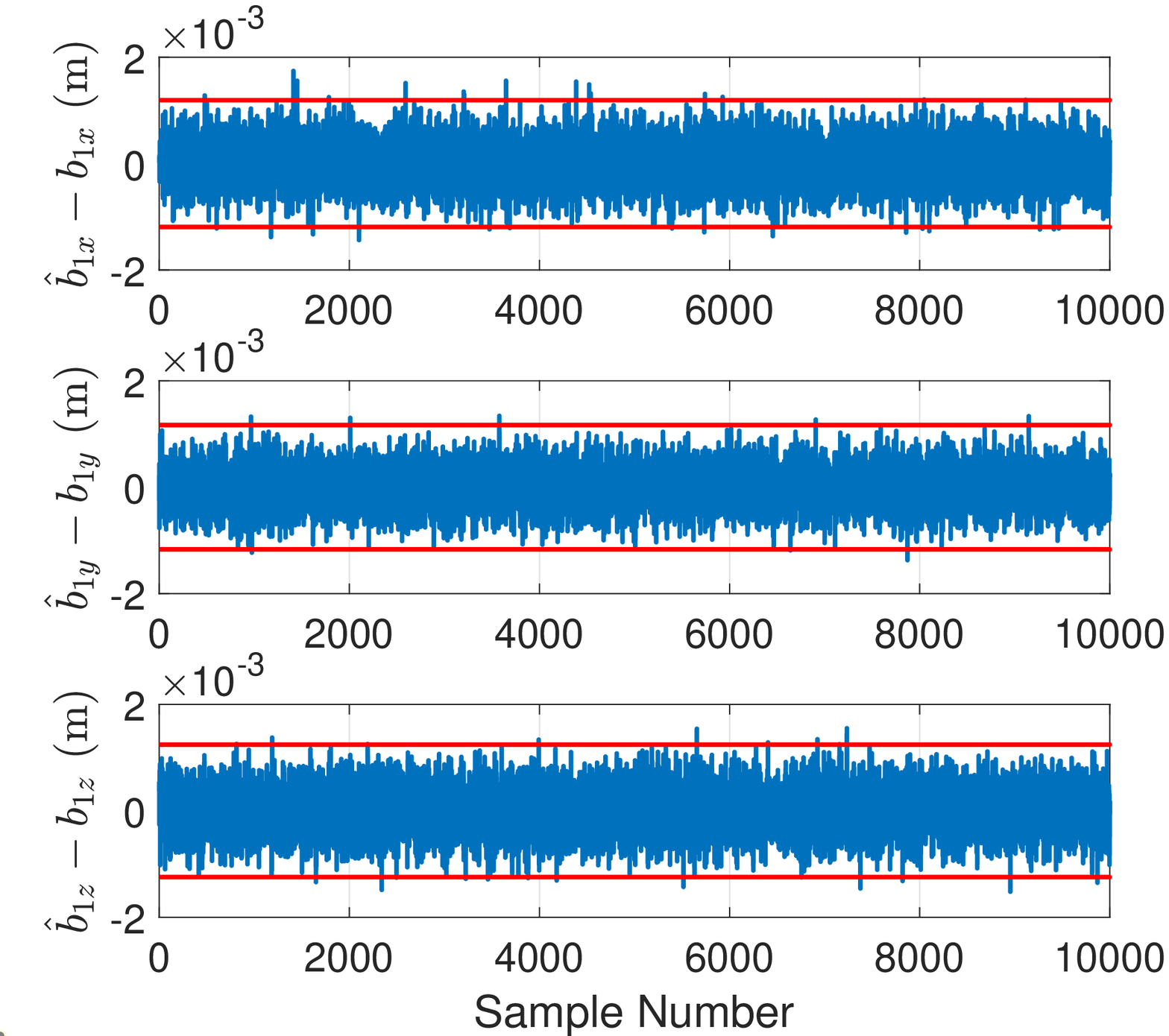}\label{fig:b_est}}
            \subfigure[{\bf Residual Errors}]
      {\includegraphics[width=0.49\textwidth]{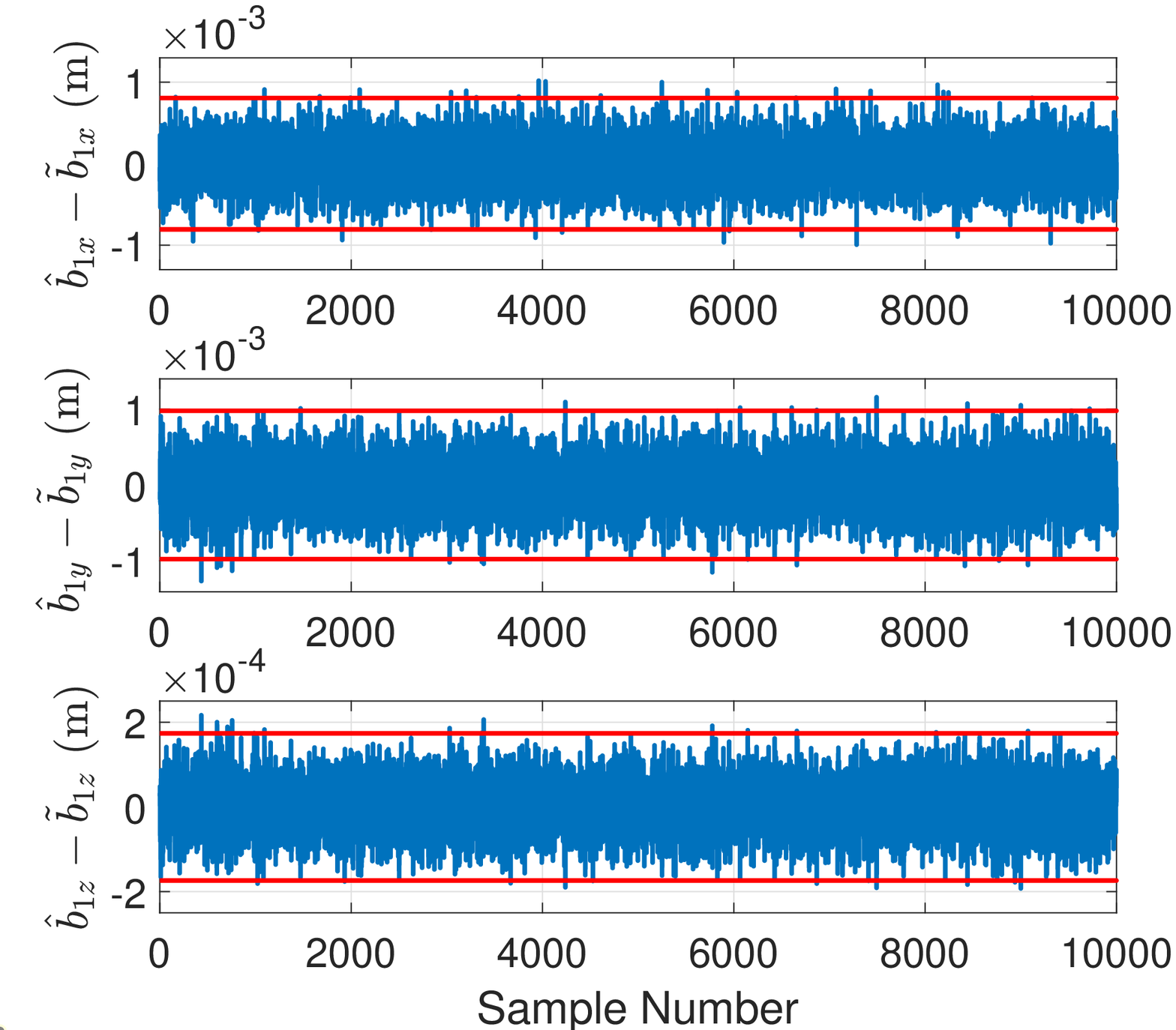}\label{fig:b_res}}
    \caption{Monte Carlo simulation for estimates and residuals of the vector observation $\mathbf{b}_1$.}
  \end{centering}
\end{figure}

\begin{figure}[h!]
  \begin{centering}
      \subfigure[{\bf Estimation Errors}]
      {\includegraphics[width=0.49\textwidth]{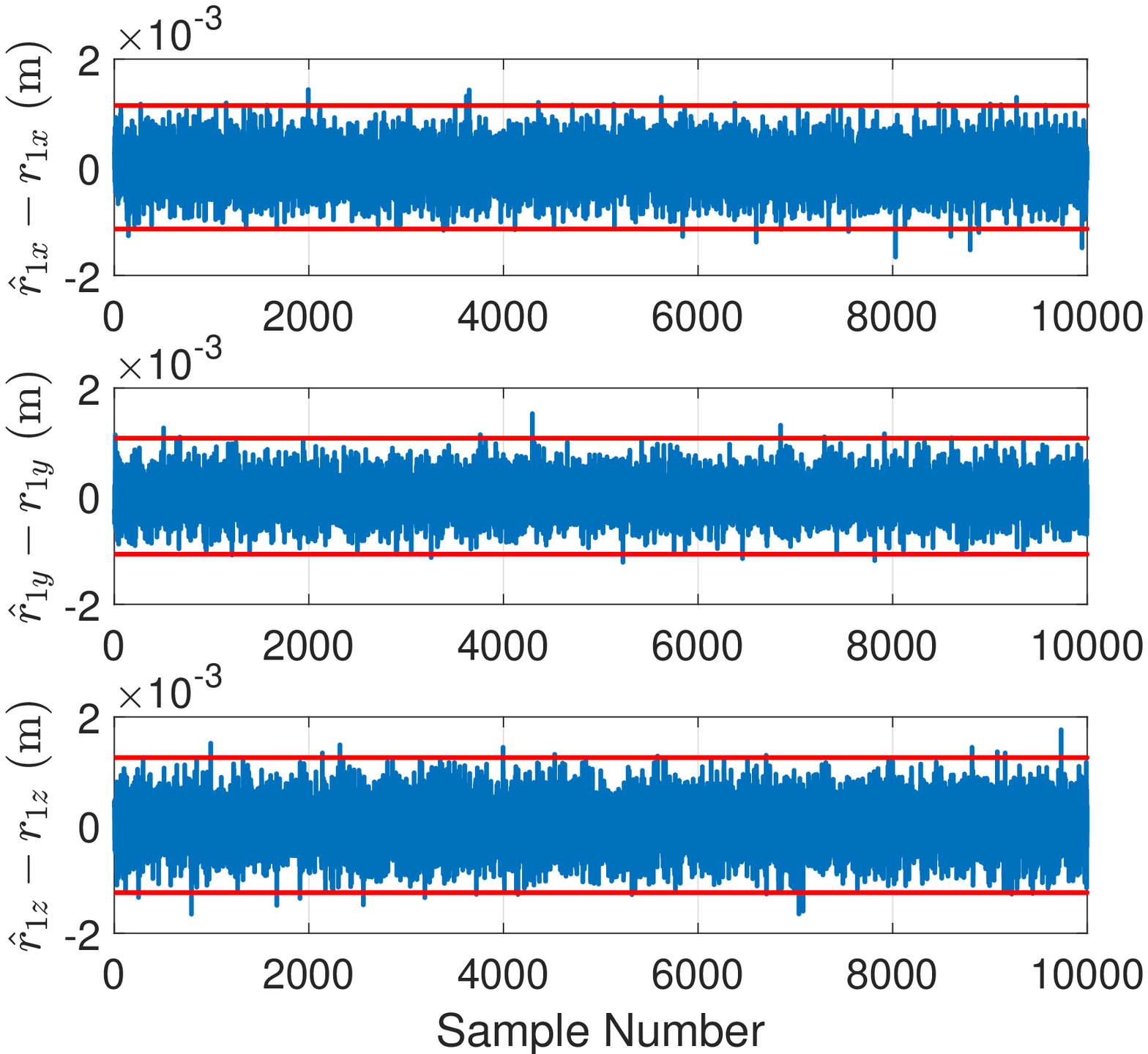}\label{fig:r_est}}
            \subfigure[{\bf Residual Errors}]
      {\includegraphics[width=0.49\textwidth]{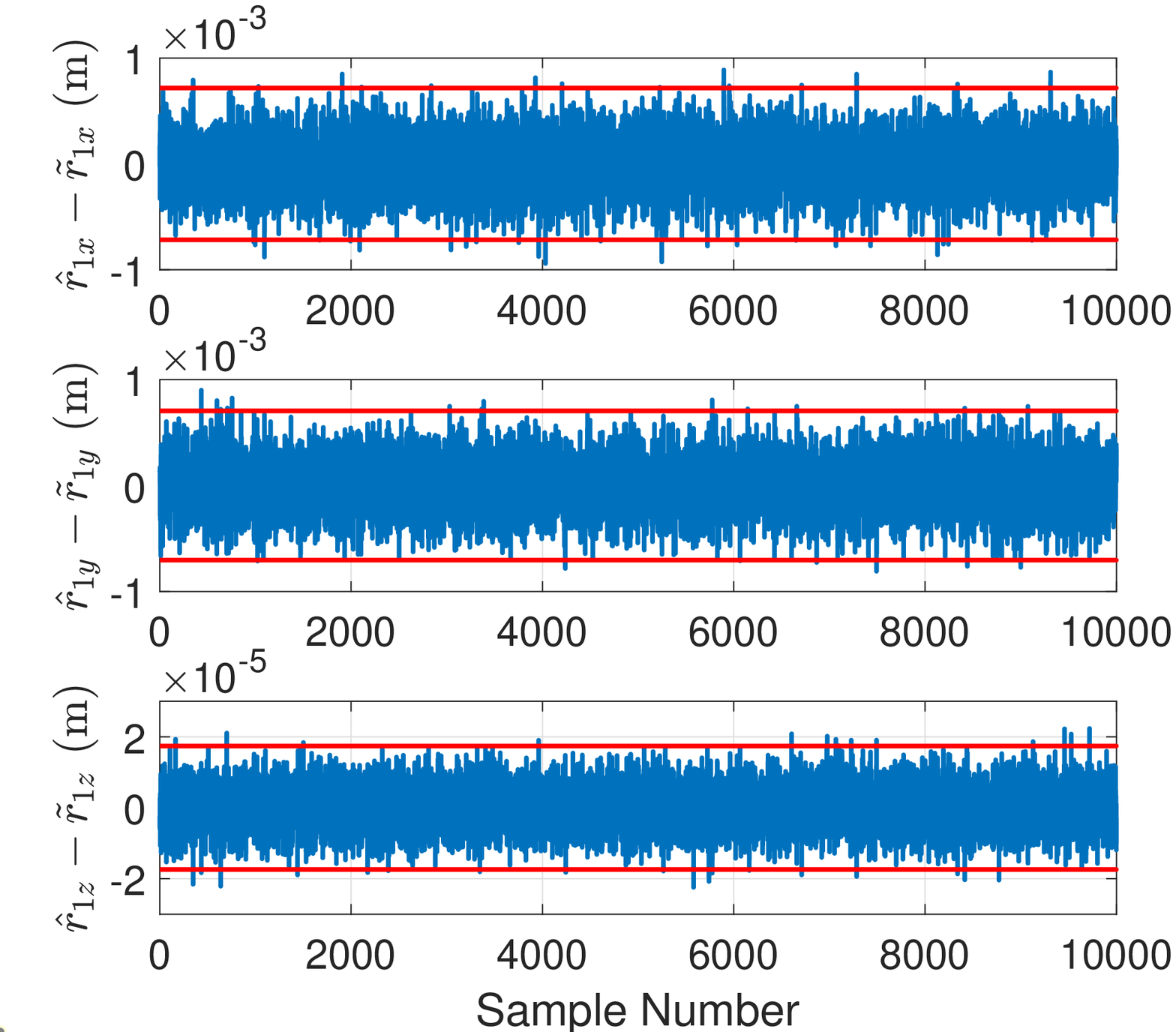}\label{fig:r_res}}
    \caption{Monte Carlo simulation for estimates and residuals of the vector observation $\mathbf{r}_1$.}
  \end{centering}
\end{figure}

A pose estimation problem is solved here with two scans of a vector-observation-enabled sensor, which can be a Lidar, with three vector observations per scan. The ground truth values for the attitude matrix $A$, translation vector $\mathbf{p}$ and vector observations $\mathbf{b}_i$, $i=1,\,2,\,3$, as well as the measurement covariances, are given as
\begin{gather*}
    A=I_{3\times3}\\
    \mathbf{p}=\begin{bmatrix}0.3 & -0.4 & 0.5\end{bmatrix}^T\\
    \mathbf{b}_1=\begin{bmatrix} 0 & 9.7590\times 10^{-2} & -1.4833 \times10^{-1}\end{bmatrix}^T\\
    \mathbf{b}_2=\begin{bmatrix} 0 & 1.9518\times 10^{-1} & -1.2855 \times10^{-2}\end{bmatrix}^T\\
    \mathbf{b}_3=\begin{bmatrix} 1 & 9.7590\times 10^{-1} & 9.8885 \times10^{-1}\end{bmatrix}^T\\
    R_1=10^{-6}\times\begin{bmatrix}0.1902 & 0.0228&   -0.0190&   -0.0345&   -0.0079&    0.0225\\
    0.0228&    0.2288&   -0.0003&    0.0145&    0.0483&   -0.0161\\
   -0.0190&   -0.0003&    0.3554&    0.0765&   -0.0180&    0.1386\\
   -0.0345&    0.0145&    0.0765&    0.2566&   -0.0201&    0.0408\\
   -0.0079&    0.0483&   -0.0180&   -0.0201 &   0.2621&   -0.0800\\
    0.0225&   -0.0161&    0.1386&    0.0408&   -0.0800&    0.3349\end{bmatrix}\\
    R_2=10^{-6}\times\begin{bmatrix}0.1981&    0.0213&    0.0021&   -0.0519&   -0.0218&   -0.0231\\
    0.0213&    0.1980&   -0.0264&    0.0023&   -0.0116&    0.0030\\
    0.0021&   -0.0264&    0.2040&   -0.0456 &   0.0273&   -0.0152\\
   -0.0519 &   0.0023 &  -0.0456&    0.2481&    0.0025&    0.0258\\
   -0.0218&   -0.0116&    0.0273&    0.0025&    0.1933&    0.0069\\
   -0.0231&    0.0030&   -0.0152&    0.0258&    0.0069&    0.1851\end{bmatrix}\\
   R_3=10^{-6}\begin{bmatrix}0.1705&   -0.0071&   -0.0154&   -0.0247&    0.0081&    0.0049\\
   -0.0071&    0.2036&    0.0038&    0.0259&   -0.0311&    0.0064\\
   -0.0154 &   0.0038 &   0.1910&    0.0376&    0.0085&    0.0166\\
   -0.0247&    0.0259&    0.0376&    0.2738&   -0.0153&    0.0170\\
    0.0081 &  -0.0311 &   0.0085&   -0.0153&    0.1850&   -0.0114\\
    0.0049&    0.0064&    0.0166&    0.0170&   -0.0114&    0.2049\end{bmatrix}
\end{gather*}
The true observation vectors $\mathbf{r}_i$ are generated by the constraint in Eq.~\eqref{eqn:det_tls_sensor_model}.
Both measurement and actual observation vectors have a meter unit.
A Monte-Carlo simulation with 10,000 samples is performed here to showcase how well the $3\sigma$ bounds generated by the covariance expressions in Eqs.~\eqref{eqn:P_delta_alpha_final} and \eqref{calc_cov_phat} for the attitude and translation vectors, respectively, Eqs.~\eqref{eqn:P_bi} and \eqref{eqn:P_ri} for estimated observations, and Eqs.~\eqref{eqn:cov_bi} and \eqref{eqn:cov_ri} for residuals of observations are bounding their corresponding estimate errors or residuals.
Artificial noise is generated from a Gaussian distribution with zero mean and covariance of $R_{r_i}$, $R_{b_i}$ and $R_{rb_i}$ to produce $\tilde{\mathbf{r}}_i$ and $\tilde{\mathbf{b}}_i$ samples.
The covariance matrices of the measurements $R_{r_i}$, $R_{b_i}$ and $R_{rb_i}$ are generated randomly with a signal-to-noise ratio of around $10^{-4}$. Note that by default, the measurement covariance is selected to be positive-definite while being random.
Singularities in the measurement covariance matrix can be handled if they exist, but this is not the focus of this paper.
Figure \ref{fig:rpy} shows the attitude errors in terms of roll, pitch and yaw angles in degrees from the Monte-Carlo samples. The blue line depicts the estimation errors, and the red lines are the $3\sigma$ bounds computed from the estimate error-covariances. Figure \ref{fig:p} shows the translation vector estimate error in the $x$, $y$ and $z$ directions, respectively.
It is seen that the estimate errors are well-bounded by their corresponding $3\sigma$ bounds. Figure \ref{fig:b_est} shows the estimation errors, and Figure \ref{fig:b_res} shows the residual errors, for observation vector $\mathbf{b}_1$.
Figures \ref{fig:r_est} and \ref{fig:r_res} show the same results for the observation vector $\mathbf{r}_1$, respectively.
It is seen that observation vectors are also bounded by their corresponding $3\sigma$ bounds, provided by the covariances of estimates as well as the residuals. 

\section{Conclusions}
This paper develops an analytical framework for an efficient pose estimation problem within the first-order approximation of estimate errors. Pose estimation is central to simultaneous localization and mapping (SLAM) problems. Efficient estimation makes the controller policy easier to track the desired signals to obtain a more accurate estimate of the states, which in turn consumes lower energy levels in the control action. The static SLAM problem is shown to be solved as a total least squares (TLS) problem. A quadratic cost function based on the TLS formulation is introduced for taking into account the attitude matrix and translation vector. The weight matrix in the cost function is extracted from the most generic positive-definite fully populated matrix to include the correlations between the observation vectors in the most general case. This work expands previously derived pose estimation solutions where an isotropic assumption is applied to the measurement noise covariance. The cost function is then written in the attitude-only format. The covariance expression of the attitude error is provided within the small-angle assumption. This assumption leads to a second-order approximation of the cost function in terms of the attitude error. Covariance expressions for the translation vector as well as the estimates and residuals of the observation vectors are obtained analytically. The Fisher information matrix (FIM) is derived, and the covariance expressions are proven to be the block inverses of FIM, proving the equality in the Cram\'er-Rao lower bound and thus the efficiency of estimates within the first-order attitude errors. A simulation framework showcases the efficacy of the covariance analyses by simulating observation vectors in a pose estimation problem with 10,000 Monte-Carlo samples. Also, an advantage of the proposed solution is that it provides an estimate of the residual for the observation vectors as well as their covariance. These quantities are useful to assess the performance of the pose estimation solution when no access to the ground truth data in an actual SLAM application is available.

\section*{Appendix}
A list of some preliminary equations used in the paper is shown here.
For the cross product of two $3\times 1$ vectors $\mathbf{a}$ and $\mathbf{b}$, the following relation is given:
\begin{align}
    \mathbf{a}\times\mathbf{b}=[\mathbf{a}\times]\mathbf{b}=-\mathbf{b}\times\mathbf{a}=-[\mathbf{b}\times]\mathbf{a}\label{eqn:cross_prod_def}
\end{align}
The cross product matrix of a vector $\mathbf{a}$ is a skew-symmetric matrix, so that
\begin{align}\label{eqn:skew-symmetric_cross_prod}
    [\mathbf{a}\times]^T=-[\mathbf{a}\times]
\end{align}
where $[\mathbf{a}\times]$ denotes the $3\times3$ cross product matrix constructed from the vector $\mathbf{a}$ as
\begin{align}
    [\mathbf{a}\times]=\begin{bmatrix}0 & -a_3 & a_2\\
    a_3 & 0 & -a_1\\
    -a_2 & a_1 & 0\end{bmatrix}
\end{align}
The Kronecker product \cite{steeb2011matrix} also can be combined with the vec operator in the following identity, where the
notation vec(.) \cite{1084534} for an $m\times n$ matrix $A$ results in an $(mn)\times 1$ matrix vec$(A)$ that is constructed by stacking the columns of $A$:
\begin{equation}\label{eqn:kronecker_vec}
    A \mathbf{z}=(\mathbf{z}^T \otimes I_{m\times m})\text{vec}(A)
\end{equation}

\bibliography{sample}

\begin{thebibliography}{38}
\newcommand{\enquote}[1]{``#1''}
\providecommand{\natexlab}[1]{#1}
\providecommand{\url}[1]{\texttt{#1}}
\providecommand{\urlprefix}{URL }
\expandafter\ifx\csname urlstyle\endcsname\relax
  \providecommand{\doi}[1]{\discretionary{}{}{}https://doi.org/#1}\else
  \providecommand{\doi}[1]{\discretionary{}{}{}\urlstyle{rm}\url{https://doi.org/#1}}\fi

\bibitem[{Crassidis et~al.(2007)Crassidis, Markley, and
  Cheng}]{survey_nonlin_att}
Crassidis, J.~L., Markley, F.~L., and Cheng, Y., \enquote{Survey of Nonlinear
  Attitude Estimation Methods,} \emph{Journal of Guidance, Control, and
  Dynamics}, Vol.~30, No.~1, 2007, pp. 12--28.
\newblock \doi{10.2514/1.22452}.

\bibitem[{Wahba(1965)}]{wahba1965least}
Wahba, G., \enquote{A Least Squares Estimate of Satellite Attitude,} \emph{SIAM
  review}, Vol.~7, No.~3, 1965, pp. 409--409.
\newblock \doi{10.1137/1008080}.

\bibitem[{Markley and Mortari(2000)}]{markley2000quaternion}
Markley, F.~L., and Mortari, D., \enquote{Quaternion Attitude Estimation Using
  Vector Observations,} \emph{The Journal of the Astronautical Sciences},
  Vol.~48, No.~2, 2000, pp. 359--380.
\newblock \doi{10.1007/BF03546284}.

\bibitem[{Eggert et~al.(1997)Eggert, Lorusso, and
  Fisher}]{eggert1997estimating}
Eggert, D.~W., Lorusso, A., and Fisher, R.~B., \enquote{Estimating 3-D Rigid
  Body Transformations: A Comparison of Four Major Algorithms,} \emph{Machine
  vision and applications}, Vol.~9, No.~5, 1997, pp. 272--290.
\newblock \doi{10.1007/s001380050048}.

\bibitem[{Erol et~al.(2007)Erol, Bebis, Nicolescu, Boyle, and
  Twombly}]{ErolAli2007Vhpe}
Erol, A., Bebis, G., Nicolescu, M., Boyle, R.~D., and Twombly, X.,
  \enquote{Vision-Based Hand Pose Estimation: A Review,} \emph{Computer vision
  and image understanding}, Vol. 108, No.~1, 2007, pp. 52--73.
\newblock \doi{10.1016/j.cviu.2006.10.012}.

\bibitem[{Kelsey et~al.(2006)Kelsey, Byrne, Cosgrove, Seereeram, and
  Mehra}]{kelsey2006vision}
Kelsey, J.~M., Byrne, J., Cosgrove, M., Seereeram, S., and Mehra, R.~K.,
  \enquote{Vision-Based Relative Pose Estimation for Autonomous Rendezvous and
  Docking,} \emph{2006 IEEE aerospace conference}, IEEE, 2006, pp. 20--pp.
\newblock \doi{10.1109/AERO.2006.1655916}.

\bibitem[{Jurie et~al.(2002)Jurie, Dhome et~al.}]{jurie2002real}
Jurie, F., Dhome, M., et~al., \enquote{Real Time Robust Template Matching,}
  \emph{BMVC}, Vol. 2002, 2002, pp. 123--132.
\newblock \doi{10.5244/C.16.10}.

\bibitem[{Isard and Blake(1998)}]{isard1998condensation}
Isard, M., and Blake, A., \enquote{Condensation—Conditional Density
  Propagation for Visual Tracking,} \emph{International journal of computer
  vision}, Vol.~29, No.~1, 1998, pp. 5--28.
\newblock \doi{10.1023/A:1008078328650}.

\bibitem[{Schonberger and Frahm(2016)}]{schonberger2016structure}
Schonberger, J.~L., and Frahm, J.-M., \enquote{Structure-From-Motion
  Revisited,} \emph{Proceedings of the IEEE conference on computer vision and
  pattern recognition}, 2016, pp. 4104--4113.
\newblock \doi{10.1109/CVPR.2016.445}.

\bibitem[{Melekhov et~al.(2017)Melekhov, Ylioinas, Kannala, and
  Rahtu}]{melekhov2017relative}
Melekhov, I., Ylioinas, J., Kannala, J., and Rahtu, E., \enquote{Relative
  Camera Pose Estimation Using Convolutional Neural Networks,}
  \emph{International Conference on Advanced Concepts for Intelligent Vision
  Systems}, Springer, 2017, pp. 675--687.
\newblock \doi{10.1007/978-3-319-70353-4_57}.

\bibitem[{Li et~al.(2018)Li, Wang, Ji, Xiang, and Fox}]{li2018deepim}
Li, Y., Wang, G., Ji, X., Xiang, Y., and Fox, D., \enquote{Deepim: Deep
  Iterative Matching for 6d Pose Estimation,} \emph{Proceedings of the European
  Conference on Computer Vision (ECCV)}, 2018, pp. 683--698.
\newblock \doi{10.1007/s11263-019-01250-9}.

\bibitem[{Toshev and Szegedy(2014)}]{toshev2014deeppose}
Toshev, A., and Szegedy, C., \enquote{Deeppose: Human Pose Estimation via Deep
  Neural Networks,} \emph{Proceedings of the IEEE conference on computer vision
  and pattern recognition}, 2014, pp. 1653--1660.
\newblock \doi{10.1109/CVPR.2014.214}.

\bibitem[{Rusu and Cousins(2011)}]{rusu20113d}
Rusu, R.~B., and Cousins, S., \enquote{3d is Here: Point Cloud Library (pcl),}
  \emph{2011 IEEE international conference on robotics and automation}, IEEE,
  2011, pp. 1--4.
\newblock \doi{10.1109/ICRA.2011.5980567}.

\bibitem[{Opromolla et~al.(2015)Opromolla, Fasano, Rufino, and
  Grassi}]{opromolla2015uncooperative}
Opromolla, R., Fasano, G., Rufino, G., and Grassi, M., \enquote{Uncooperative
  Pose Estimation with a LIDAR-Based System,} \emph{Acta Astronautica}, Vol.
  110, 2015, pp. 287--297.
\newblock \doi{10.1016/j.actaastro.2014.11.003}.

\bibitem[{Rusinkiewicz and Levoy(2001)}]{rusinkiewicz2001efficient}
Rusinkiewicz, S., and Levoy, M., \enquote{Efficient Variants of the ICP
  Algorithm,} \emph{Proceedings third international conference on 3-D digital
  imaging and modeling}, IEEE, 2001, pp. 145--152.
\newblock \doi{10.1109/IM.2001.924423}.

\bibitem[{Woo et~al.(2002)Woo, Kang, Wang, and Lee}]{woo2002new}
Woo, H., Kang, E., Wang, S., and Lee, K.~H., \enquote{A New Segmentation Method
  for Point Cloud Data,} \emph{International Journal of Machine Tools and
  Manufacture}, Vol.~42, No.~2, 2002, pp. 167--178.
\newblock \doi{10.1016/S0890-6955(01)00120-1}.

\bibitem[{Trevor et~al.(2013)Trevor, Gedikli, Rusu, and
  Christensen}]{trevor2013efficient}
Trevor, A.~J., Gedikli, S., Rusu, R.~B., and Christensen, H.~I.,
  \enquote{Efficient Organized Point Cloud Segmentation with Connected
  Components,} \emph{Semantic Perception Mapping and Exploration (SPME)}, 2013.

\bibitem[{Landrieu and Simonovsky(2018)}]{landrieu2018large}
Landrieu, L., and Simonovsky, M., \enquote{Large-Scale Point Cloud Semantic
  Segmentation with Superpoint Graphs,} \emph{Proceedings of the IEEE
  Conference on Computer Vision and Pattern Recognition}, 2018, pp. 4558--4567.
\newblock \doi{10.1109/CVPR.2018.00479}.

\bibitem[{Lindeberg(2012)}]{lindeberg2012scale}
Lindeberg, T., \enquote{Scale Invariant Feature Transform,} 2012.
\newblock \doi{10.4249/scholarpedia.10491}.

\bibitem[{Bay et~al.(2006)Bay, Tuytelaars, and Van~Gool}]{bay2006surf}
Bay, H., Tuytelaars, T., and Van~Gool, L., \enquote{Surf: Speeded Up Robust
  Features,} \emph{European conference on computer vision}, Springer, 2006, pp.
  404--417.
\newblock \doi{10.1016/j.cviu.2007.09.014}.

\bibitem[{Rusu et~al.(2008)Rusu, Blodow, Marton, and Beetz}]{rusu2008aligning}
Rusu, R.~B., Blodow, N., Marton, Z.~C., and Beetz, M., \enquote{Aligning Point
  Cloud Views Using Persistent Feature Histograms,} \emph{2008 IEEE/RSJ
  international conference on intelligent robots and systems}, IEEE, 2008, pp.
  3384--3391.
\newblock \doi{10.1109/IROS.2008.4650967}.

\bibitem[{Serafin and Grisetti(2015)}]{serafin2015nicp}
Serafin, J., and Grisetti, G., \enquote{NICP: Dense Normal Based Point Cloud
  Registration,} \emph{2015 IEEE/RSJ International Conference on Intelligent
  Robots and Systems (IROS)}, IEEE, 2015, pp. 742--749.
\newblock \doi{10.1109/IROS.2015.7353455}.

\bibitem[{Hashim(2020)}]{hashim2020attitude}
Hashim, H.~A., \enquote{Attitude Determination and Estimation Using Vector
  Observations: Review, Challenges and Comparative Results,} \emph{arXiv
  preprint arXiv:2001.03787}, 2020.

\bibitem[{Crassidis(2011)}]{alma9938879548204803}
Crassidis, J.~L., \emph{Optimal estimation of dynamic systems John L.
  Crassidis, John L. Junkins.}, 2\textsuperscript{nd} ed., Chapman \& Hall/CRC
  applied mathematics and nonlinear science series, Chapman and Hall/CRC, Boca
  Raton, Fla, 2011, Chap.~2.
\newblock \doi{10.1201/b11154}.

\bibitem[{Cheng and Crassidis(2021)}]{cheng2021optimal}
Cheng, Y., and Crassidis, J.~L., \enquote{Optimal Pose Estimation with
  Error-Covariance Analysis,} \emph{AIAA Scitech 2021 Forum}, 2021, p. 1758.
\newblock \doi{10.2514/6.2021-1758}.

\bibitem[{Est{\'e}par et~al.(2004)Est{\'e}par, Brun, and
  Westin}]{estepar2004robust}
Est{\'e}par, R. S.~J., Brun, A., and Westin, C.-F., \enquote{Robust Generalized
  Total Least Squares Iterative Closest Point Registration,}
  \emph{International Conference on Medical Image Computing and
  Computer-Assisted Intervention}, Springer, 2004, pp. 234--241.
\newblock \doi{10.1007/978-3-540-30135-6_29}.

\bibitem[{Shuster(1993)}]{shuster1993survey}
Shuster, M.~D., \enquote{{A Survey of Attitude Representations},} \emph{The
  Journal of the Astronautical Sciences}, Vol.~41, No.~4, 1993, pp. 439--517.

\bibitem[{Golub(1973)}]{golub1973some}
Golub, G.~H., \enquote{Some Modified Matrix Eigenvalue Problems,} \emph{Siam
  Review}, Vol.~15, No.~2, 1973, pp. 318--334.
\newblock \doi{10.1137/1015032}.

\bibitem[{Markovsky and Van~Huffel(2007)}]{markovsky2007overview}
Markovsky, I., and Van~Huffel, S., \enquote{Overview of Total Least-Squares
  Methods,} \emph{Signal processing}, Vol.~87, No.~10, 2007, pp. 2283--2302.
\newblock \doi{10.1016/j.sigpro.2007.04.004}.

\bibitem[{Golub and Van~Loan(1980)}]{golub1980analysis}
Golub, G.~H., and Van~Loan, C.~F., \enquote{An Analysis of the Total Least
  Squares Problem,} \emph{SIAM journal on numerical analysis}, Vol.~17, No.~6,
  1980, pp. 883--893.
\newblock \doi{10.1137/0717073}.

\bibitem[{Cram{\'e}r(1999)}]{cramer1999mathematical}
Cram{\'e}r, H., \emph{Mathematical Methods of Statistics}, Vol.~43, Princeton
  university press, 1999.
\newblock \doi{10.1515/9781400883868}.

\bibitem[{Crassidis and Cheng(2019)}]{crassidis2019maximum}
Crassidis, J.~L., and Cheng, Y., \enquote{Maximum Likelihood Analysis of the
  Total Least Squares Problem with Correlated Errors,} \emph{Journal of
  Guidance, Control, and Dynamics}, Vol.~42, No.~6, 2019, pp. 1204--1217.
\newblock \doi{10.2514/6.2019-1931}.

\bibitem[{Steeb and Hardy(2011)}]{steeb2011matrix}
Steeb, W.-H., and Hardy, Y., \emph{Matrix Calculus and Kronecker Product: A
  Practical Approach to Linear and Multilinear Algebra}, World Scientific
  Publishing Company, 2011.
\newblock \doi{10.1142/8030}.

\bibitem[{Shuster and Oh(1981)}]{shuster1981three}
Shuster, M.~D., and Oh, S.~D., \enquote{Three-Axis Attitude Determination from
  Vector Observations,} \emph{Journal of guidance and Control}, Vol.~4, No.~1,
  1981, pp. 70--77.
\newblock \doi{10.2514/3.19717}.

\bibitem[{Amiri-Simkooei(2017)}]{amiri2017weighted}
Amiri-Simkooei, A., \enquote{Weighted Total Least Squares with Singular
  Covariance Matrices subject to Weighted and Hard Constraints,} \emph{Journal
  of Surveying Engineering}, Vol. 143, No.~4, 2017, p. 04017018.
\newblock \doi{10.1061/(ASCE)SU.1943-5428.0000239}.

\bibitem[{Markley and Crassidis(2014)}]{markley2014fundamentals}
Markley, F.~L., and Crassidis, J.~L., \emph{Fundamentals of Spacecraft Attitude
  Determination and Control}, Springer, 2014.

\bibitem[{Sherman and Morrison(1950)}]{sherman1950adjustment}
Sherman, J., and Morrison, W.~J., \enquote{Adjustment of an Inverse Matrix
  Corresponding to a Change in One Element of a Given Matrix,} \emph{The Annals
  of Mathematical Statistics}, Vol.~21, No.~1, 1950, pp. 124--127.
\newblock \doi{10.1214/aoms/1177729893}.

\bibitem[{Brewer(1978)}]{1084534}
Brewer, J., \enquote{Kronecker Products and Matrix Calculus in System Theory,}
  \emph{IEEE Transactions on Circuits and Systems}, Vol.~25, No.~9, 1978, pp.
  772--781.
\newblock \doi{10.1109/TCS.1978.1084534}.

\end{thebibliography}

\end{document}